%% file: main_arxiv.tex
\documentclass[twoside,11pt]{article}

\usepackage[T1]{fontenc}
\usepackage[preprint]{arxiv}
\usepackage{microtype}
\usepackage{amsmath}
\usepackage{booktabs}
\usepackage{tabularx}
\usepackage[table]{xcolor}
\usepackage{colortbl}
\definecolor{darkgreen}{rgb}{0.0, 0.5, 0.0}
\definecolor{darkred}{rgb}{0.6, 0.0, 0.0}
\definecolor{gray}{rgb}{0.5, 0.5, 0.5}
\definecolor{myblue}{RGB}{0, 0, 180}
\definecolor{mygreen}{RGB}{0, 128, 0}
\definecolor{myred}{RGB}{196, 0, 0}
\definecolor{better}{RGB}{204,255,204}
\definecolor{worse}{RGB}{255,204,203}
\definecolor{neutral}{RGB}{255,255,255}
\usepackage{listings}
\usepackage{placeins}
\usepackage{url}
\usepackage{lastpage}

\providecommand{\cref}[1]{\ref{#1}}
\providecommand{\Cref}[1]{\ref{#1}}
\input{acronyms.tex}
\let\origgls\gls
\let\origacrshort\acrshort
\let\origacrlong\acrlong
\renewcommand{\gls}[1]{\ifglsentryexists{#1}{\origgls{#1}}{\texttt{#1}}}
\renewcommand{\acrshort}[1]{\ifglsentryexists{#1}{\origacrshort{#1}}{\texttt{#1}}}
\renewcommand{\acrlong}[1]{\ifglsentryexists{#1}{\origacrlong{#1}}{\texttt{#1}}}
\input{notation.tex}

\usepackage[scaled=0.85]{DejaVuSansMono}
\definecolor{codegreen}{rgb}{0,0.6,0}
\definecolor{codegray}{rgb}{0.5,0.5,0.5}
\definecolor{codepurple}{rgb}{0.58,0,0.82}
\definecolor{backcolour}{rgb}{0.95,0.95,0.95}

\lstdefinestyle{tgpcode}{
    backgroundcolor=\color{backcolour},   
    commentstyle=\color{codegreen},
    keywordstyle=\color{magenta},
    numberstyle=\tiny\color{codegray},
    stringstyle=\color{codepurple},
    identifierstyle=\color{black},
    basicstyle=\ttfamily\footnotesize,
    rulecolor=\color{gray},
    frameround=tttt,
    frame=single,
    xleftmargin=14pt,
    xrightmargin=4pt,
    breakatwhitespace=false,         
    breaklines=true,                 
    captionpos=b,                    
    keepspaces=true,                 
    numbers=left,                    
    numbersep=8pt,                  
    showspaces=false,                
    showstringspaces=false,
    showtabs=false,                  
    tabsize=2
}
\jmlrheading{0}{2026}{1-\pageref{LastPage}}{}{}{}{Carlo Abate and Ivan Marisca and Filippo Maria Bianchi}
\ShortHeadings{Torch Geometric Pool}{Abate, Marisca, and Bianchi}
\firstpageno{1}

\title{Torch Geometric Pool: the PyTorch library for pooling in Graph Neural Networks}

\author{\name Carlo Abate \email carlo.abate@uit.no \\
       \addr UiT the Arctic University of Norway
       \AND
       \name Ivan Marisca \email ivan.marisca@usi.ch \\
       \addr Università della Svizzera italiana, Switzerland
       \AND
       \name Filippo Maria Bianchi \email filippo.m.bianchi@uit.no \\
       \addr UiT the Arctic University of Norway\\
       NORCE Norwegian Research Centre AS}

\begin{document}

\maketitle

\begin{abstract}
       Torch Geometric Pool (\texttt{tgp}) is a pooling library built on top of PyTorch Geometric.
       Graph pooling methods differ in how they assign nodes to supernodes, how they handle batches, what they return after pooling, and whether they expose auxiliary losses.
       These differences make it hard to compare methods or reuse the same model code across them.
       \texttt{tgp} addresses this problem with a common software interface based on the Select-Reduce-Connect-Lift (SRCL) decomposition.
       The library provides 20 hierarchical poolers, standardized output objects, standalone readout modules, support for dense poolers in batched and unbatched mode, and workflows for caching and pre-coarsening.
       It is released under the MIT license on GitHub and PyPI, with comprehensive documentation, tutorials, and examples.
\end{abstract}

\begin{keywords}
       graph neural networks, graph pooling, PyTorch Geometric, scientific software
\end{keywords}

\section{Introduction}

Graph pooling builds a smaller graph from a larger one.
In a hierarchical graph neural network, pooling lets the model move to a coarser representation by combining information from parts of the original graph, and then continue message passing or readout at that new scale \citep{grattarola2022understanding}.
Pooling methods are now numerous but still cumbersome to use.
They differ in what they compute, what tensors they return, how they handle batches, and whether they use auxiliary losses.
While those differences matter for the method, they should not force rewriting the model around it.
General-purpose graph learning libraries provide the surrounding infrastructure, and some of them include pooling or readout modules \citep{fey2019fast,wang2019deep,grattarola2021graph}.
What is still missing is a software focused on pooling, so that methods are easier to compare, reuse, and extend.
Without a common API, swapping poolers requires changing the forward passes, loss handling logic, readout code, and logging routines.

\texttt{tgp} is built on top of \gls{pyg} and treats pooling as a first-class software abstraction.
It relies on the Select-Reduce-Connect-Lift (SRCL) framework \citep{grattarola2022understanding} as the common structure to unify all methods.
\texttt{tgp} exposes a single construction entry point, \texttt{get\_pooler(name, **kwargs)};
poolers share one base interface, and downstream code consumes one common \texttt{PoolingOutput} contract.
This paper describes the main interface of \texttt{tgp}, how it integrates into models from the \gls{pyg} framework, and the surrounding documentation and tests.
Empirical results of hierarchical graph models agree that the optimal pooling operator depends on the task, the dataset, and the rest of the model \citep{mesquita2020rethinking, grattarola2022understanding, bianchi2023expr}. 
That leaves practitioners comparing many candidates under tight engineering constraints: the training loop, loss bookkeeping, and tensor contracts should stay aligned when only the pooling layer changes.

To support that picture with numbers, we ran a large comparative study across various tasks, with \texttt{tgp} providing the universal interface required to make this efficient, reducing operator swapping to a configuration change.

\section{Software Architecture and API}
\label{sec:software}
The root abstraction in \texttt{tgp} is the \texttt{SRCPooling} base class, which implements the four-stage SRCL decomposition.
Each pooler is described through four stages: \texttt{Select}, \texttt{Reduce}, \texttt{Connect}, and \texttt{Lift} (SRCL; see Appendix~\ref{sec:background}).
\texttt{Select} decides how input nodes map to supernodes, i.e., nodes in the coarsened graph.
\texttt{Reduce} computes supernode features by aggregating the features of the nodes assigned to each supernode.
\texttt{Connect} builds the coarse graph topology.
\texttt{Lift} projects coarse features back to the original nodes when a model needs node-level outputs.
This decomposition expresses different pooling operators through a common infrastructure.

\paragraph{\texttt{SelectOutput} and \texttt{PoolingOutput}.}

In SRCL, \texttt{Select} produces a node-to-supernode assignment map, commonly represented by an assignment matrix.
\texttt{SelectOutput} stores the assignment produced by \texttt{Select}, together with metadata used by downstream stages.
The final output returned by each pooling layer is a \texttt{PoolingOutput}, which packages pooled features, pooled connectivity, batch information, the originating \texttt{SelectOutput}, and optional auxiliary losses.
Downstream code always relies on the same output contract, no matter which pooling operator is used upstream.
Field-level details are summarized in Appendix~\ref{app:api-structures}.

\paragraph{Dense and Sparse Modes.}

A dense pooler is characterized by a soft assignment, meaning that a node can contribute to multiple supernodes.
In the original dense-pooling implementations, this dense assignment was tied to a fully batched dense-tensor pipeline (padded features and dense adjacency) rather than the standard sparse \texttt{edge\_index} representation of \gls{pyg}; in \texttt{tgp}, these choices are disentangled via separate controls for execution mode (\texttt{batched}) and output format (\texttt{sparse\_output}).
\texttt{batched=True} converts sparse batch inputs into padded dense tensors.
This improves vectorization, but densifying adjacency matrices forces the materialization of zero edges.
The memory penalty compounds in batches with uneven graph sizes, where every graph must pad to match the size of the largest matrix.
With \texttt{batched=False}, dense poolers keep assignments dense while operating without padded graph tensors, evaluating objectives through sparse losses.
This avoids storing padded non-edges and is often more memory-efficient on irregular batches, at the cost of lower vectorization.

Soft-assignment poolers subclass \texttt{DenseSRCPooling}, which centralizes this execution-mode logic.
Details about execution modes, with listings, are in Appendix~\ref{app:dense-sparse-pooling}.
Dense poolers interoperate with standard sparse \gls{pyg} pipelines through \texttt{sparse\_output=True}; operator-specific support for batched and unbatched execution is summarized in Appendix~\ref{app:pooler-cheatsheet}.

\paragraph{Pre-coarsening.}

\texttt{tgp} supports offline coarsening through the dataset \texttt{pre\_transform} \texttt{PreCoarsening}, which stores a list of coarsened graph snapshots on each \texttt{Data} object before training begins.
The transform calls \texttt{multi\_level\_precoarsening} on poolers that follow the \texttt{Precoarsenable} interface.
The API is multi-level because some operators do not factorize into repeated single-level calls.
For example, \gls{sep} \citep{wu2022structural} implements its own \texttt{multi\_level\_precoarsening} so the stored hierarchy matches the method's native schedule.
During training, \texttt{PoolDataLoader} collates those precomputed levels with the original graphs into \texttt{PooledBatch} containers.
Further details are in Appendix~\ref{app:workflow-offline}.

\paragraph{Caching.}

For downstream tasks on a single graph, e.g., node classification, two caches avoid redundant work.
Pre-coarsenable poolers support \texttt{cached=True}, which memorizes the outputs of \texttt{select} and \texttt{connect} across training epochs.
Dense poolers, on the other hand, may set \texttt{cache\_preprocessing=True} to keep the dense adjacency and padded node batch produced from sparse \texttt{edge\_index} input between forward passes on the same graph.

\paragraph{Unified Readout via \texttt{AggrReduce}.}
Readout operates exactly like \texttt{Reduce}.
A standard \texttt{Reduce} stage computes node features for a pooled graph by combining subsets of original nodes into multiple supernodes.
Readout applies the same operation to combine all nodes in a graph into a single global supernode.
\gls{pyg} provides a broad family of aggregators; \texttt{AggrReduce} wraps them under this shared \texttt{Reduce} interface.
This structure allows models to swap readout behavior (for example \texttt{sum}, \texttt{mean}, \texttt{lstm}, or \texttt{set2set}) without changing the surrounding pipeline.
\texttt{GlobalReduce} implements this all-to-one mapping to produce graph-level embeddings.
Running the equivalent abstraction inside the pooler and at readout keeps tensor contracts consistent across the architecture.
Further details and a graph-classification example that uses \texttt{AggrReduce} and \texttt{GlobalReduce} is reported in Appendix~\ref{app:aggr-reduce}.

\paragraph{Compatibility Flags and Workflows.}
The default workflow is to instantiate a pooler by alias, place it between message-passing layers, and keep the rest of the model unchanged.
Poolers expose boolean flags (Table~\ref{tab:flags-workflow} in Appendix~\ref{app:workflow-custom}) that let model code choose the correct execution path without branching on specific pooler operators.
For example, a model can query \texttt{is\_precoarsenable} to decide whether the pooler's coarsening can be computed once at loading time and reused across epochs, and check \texttt{has\_loss} to know whether auxiliary losses should be added to the training objective.

Three end-to-end workflows cover the main use cases.
A \emph{graph-level} workflow uses a pooler followed by \texttt{GlobalReduce} readout for graph-level tasks (Appendix~\ref{app:workflow-graph}).
A \emph{node-level} workflow pools to a bottleneck representation and projects features back to the original nodes via \texttt{Lift} for node prediction (Appendix~\ref{app:workflow-node}).
An \emph{unsupervised clustering} workflow trains a dense pooler using only auxiliary objectives and computes partitions from the assignment matrix (Appendix~\ref{app:workflow-clustering}).

\paragraph{Modularity and Extensibility.}

The modular design of \texttt{tgp} supports extension: users can subclass \texttt{SRCPooling} to implement a new method from scratch, or reuse existing \texttt{Select}, \texttt{Reduce}, \texttt{Connect}, and \texttt{Lift} modules to assemble a new pooler.
For example, new dense trainable operators typically subclass \texttt{DenseSRCPooling}, provide a dense \texttt{Select}, and override \texttt{compute\_loss} with the new auxiliary objectives.
To add a new precoarsenable pooler, assemble it as usual from SRCL stages with a deterministic, structure-only \texttt{Select}, keep the implementation free of trainable parameters in the coarsening rule, and mix in \texttt{BasePrecoarseningMixin} so \texttt{PreCoarsening} can call it.
Finally, swapping a single stage, e.g., replacing the default sparse connector with \texttt{KronConnect}, does not require rewriting the forward pass.
Appendix~\ref{app:workflow-custom} sketches such a swap; Appendix~\ref{app:pooler-cheatsheet} lists operators and flags.

\section{Documentation and Quality}

General-purpose graph libraries such as \gls{pyg}~\citep{fey2019fast}, DGL~\citep{wang2019deep}, and Spektral~\citep{grattarola2021graph} provide broad graph-learning infrastructure and include selected pooling and readout layers.
However, pooling is not their fundamental abstraction: method-specific interfaces remain exposed, and workflows such as dense-mode configuration, auxiliary-loss handling, assignment caching, or offline pre-coarsening still have to be assembled at model level.
\texttt{tgp} complements these ecosystems by making pooling itself the root abstraction, with a unified catalog of 20 hierarchical operators and the infrastructure needed to compare and deploy them efficiently.

\texttt{tgp} is distributed through GitHub and PyPI and supports Python 3.9 to 3.12.
Optional dependency groups cover notebooks, documentation, testing, and development work.
The public repository also includes contributor instructions, a test suite, and a GitHub Actions workflow that runs linting, formatting checks, and tests with coverage.
The documentation includes installation and quick-start material, a conceptual introduction to the SRCL framework, API pages generated from docstrings, a pooler cheatsheet, executable tutorial notebooks, and example scripts for common downstream tasks.
Specifically, the tutorial notebooks cover the main software concepts from Section~\ref{sec:software}, while the repository example scripts cover the workflow families summarized in Appendix~\ref{app:workflow-listings}.

\section{Experimental comparisons and conclusions}

Identifying the optimal pooling operator for a given architecture and dataset is an inherently empirical challenge. 
To demonstrate this, we conducted an extensive experimental evaluation across unsupervised node clustering, transductive node classification, and graph-level classification. 
By training and scoring multiple poolers through the unified \texttt{tgp} pipeline, we report in the Appendix a reproducible comparative evaluation built around the same public interfaces exposed by the library.
Dataset splits, training settings, full result tables, and discussion are collected in Appendix~\ref{app:dataset-details} and \ref{app:evaluation-overview}. 
Ultimately, these results highlight a highly heterogeneous landscape where different operators dominate different tasks, confirming that no single pooling method universally excels.
This variability clearly motivates the need for \texttt{tgp}. 
When architectural choices must be driven by empirical validation, a shared interface and a uniform output contract drastically reduce the friction of comparing alternatives.
By abstracting away the complex engineering of forward passes, auxiliary losses, and memory management, \texttt{tgp} transforms the trial of new poolers into a rapid, systematic process compared to manually swapping implementations.

In conclusion, \texttt{tgp} bridges the gap between the theoretical diversity of hierarchical graph pooling and its practical realization in \gls{pyg}. 
It complements general-purpose graph libraries by elevating pooling to a first-class software abstraction, backed by specialized infrastructure for unified readout, caching, pre-coarsening, and unbatched dense execution. 
As an open-source project, we invite the community to contribute new operators, benchmarks, and ideas as the catalog continues to grow.

\input{ack}

\newpage
\bibliography{biblio}

\newpage
\input{appendix}

\end{document}

%% file: acronyms.tex
\usepackage[acronym, nohypertypes={acronym}]{glossaries}
\glsdisablehyper

\newacronym{tgp}{\texttt{tgp}}{Torch Geometric Pool}
\newacronym{pyg}{PyG}{Pytorch Geometric}
\newacronym{dgl}{DGL}{Deep Graph Library}
\newacronym{gsp}{GSP}{Graph Signal Processing}
\newacronym{src}{SRC}{Select-Reduce-Connect}
\newacronym{srcl}{SRCL}{Select-Reduce-Connect-Lift}
\newacronym{nmi}{NMI}{Normalized Mutual Information}

\newacronym{cnn}{CNN}{Convolutional Neural Network}
\newacronym{gnn}{GNN}{Graph Neural Network}
\newacronym{mlp}{MLP}{Multilayer Perceptron}
\newacronym{gcn}{GCN}{Graph Convolutional Network}
\newacronym{gin}{GIN}{Graph Isomorphism Network}

\newacronym{mp}{MP}{Message Passing}
\newacronym{ogcn}{$\omega$-GCN}{$\omega$-Graph Convolution Network}
\newacronym{hetmp}{HetMP}{Heterophilic Message Passing}

\newacronym{hosc}{HOSC}{High-Order Spectral Clustering}
\newacronym{asap}{ASAP}{Adaptive Structure Aware Pooling}
\newacronym{jbgnn}{JBPool}{Just-Balance Pooling}
\newacronym{ecpool}{ECPool}{Edge-Contraction Pooling}
\newacronym{ndp}{NDP}{Node Decimation Pooling}
\newacronym{nmf}{NMF}{Non-negative Matrix Factorization pooling}
\newglossaryentry{topk}{name=Top-$k$,description=}
\newacronym{dmon}{DMoN}{Deep Modularity Networks}
\newacronym{acc}{ACC}{Asymmetric Cheeger Cut pooling}
\newacronym{kmis}{$k$-MIS}{$k$-Maximal Independent Sets pooling}
\newglossaryentry{mincut}{name=MinCut,description=}
\newglossaryentry{maxcut}{name=MaxCut,description=}
\newacronym{tvgnn}{TVGNN}{Total Variation Graph Neural Network}
\newacronym{pan}{PAN}{Path integral based pooling}
\newacronym{sag}{SAG}{Self-Attention Graph pooling}
\newacronym{bnpool}{BNPool}{Bayesian Nonparametric Pooling}
\newglossaryentry{graclus}{name=GraClus,description=}
\newacronym{lapool}{LaPool}{Laplacian Pooling}
\newacronym{sep}{SEP}{Structural Entropy Pooling}
\newglossaryentry{eigen}{name=EigenPool,description=}
\newglossaryentry{diff}{name=DiffPool,description=}
\newglossaryentry{nopool}{name=NoPool,description=}

\newacronym{ogb}{ogbg-ppa}{protein-protein association networks}
\newacronym{gcb}{GCB-H}{Graph Classification Benchmark Hard}
\newacronym{lrgb}{LRGB}{Long-Range Graph Benchmark}

%% file: notation.tex
\usepackage{amsmath, amsfonts, bm}

\def\1{\bm{1}}

\def\mA{{\bm{A}}}

\def\mS{{\bm{S}}}

\def\mX{{\bm{X}}}

\DeclareMathAlphabet{\mathsfit}{\encodingdefault}{\sfdefault}{m}{sl}
\SetMathAlphabet{\mathsfit}{bold}{\encodingdefault}{\sfdefault}{bx}{n}

\def\eqref#1{equation~\ref{#1}}



%% file: ack.tex
\section*{Acknowledgments}
This work is supported by the Norwegian Research Council project no. 345017~(\textit{RELAY:~Relational Deep Learning for Energy Analytics}).
The authors wish to thank Nvidia Corporation for donating some of the GPUs used in this project.

%% file: appendix.tex
\appendix

\section{SRCL: a unified graph pooling framework} \label{sec:background}

\begin{figure}[htbp]
    \centering
    \includegraphics[width=0.45\linewidth]{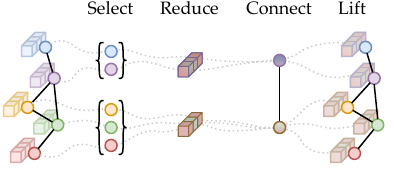}
    \caption{Overview of \gls{srcl}.
    The first three SRCL stages coarsen the graph by mapping nodes to supernodes. 
    \texttt{Lift} is the inverse operation that reprojects supernode features back to the original node space.}
    \label{fig:srcl}
\end{figure}

While there are profound differences between existing graph pooling operators, most of them can be expressed through the \gls{srcl} framework \citep{grattarola2022understanding}.
Specifically, a pooling operator \texttt{POOL}: $(\mA,\mX) \mapsto (\mA', \mX')$ can be expressed as the combination of the following sub-operators, as illustrated in Figure~\ref{fig:srcl}:

\begin{itemize}
    \item \textbf{\texttt{Select}}: $(\mA,\mX) \mapsto \mS \in \mathbb{R}^{N \times K}$, defines how the $N$ original nodes are mapped to the $K$ pooled nodes, called \emph{supernodes}. The output $\mS$ is the \emph{selection matrix} (or assignment matrix).

    \item \textbf{\texttt{Reduce}}: $(\mX, \mS) \mapsto \mX' \in \mathbb{R}^{K \times F}$, yields the features of the supernodes based on the original features and the selection matrix. For example, $\mX' = \mS^\top\mX$.

    \item \textbf{\texttt{Connect}}: $(\mA, \mS) \mapsto \mA' \in \mathbb{R}^{K \times K}_{\geq0}$, generates the new adjacency matrix for the coarsened graph, by merging the edges of the nodes mapped to the same supernode, e.g., $\mA' = \mS^\top\mA\mS$.
\end{itemize}

Furthermore, to support node-level tasks, this framework is often complemented by a fourth operation: 

\begin{itemize}
    \item \textbf{\texttt{Lift}}: $(\mX', \mS) \mapsto \mX_{\text{lift}} \in \mathbb{R}^{N \times F'}$, projects the features of the supernodes back to the original $N$ nodes, e.g., $\mX_{\text{lift}} = \mS\mX'$.
\end{itemize}

\section{Standardized Data Structures}
\label{app:api-structures}

The modularity and extensibility of \gls{tgp} depend on two shared data structures that define the contracts between pooling stages.


\paragraph{\texttt{SelectOutput}} is the standard output of every \texttt{Select} module. 
Its central field is the assignment matrix \texttt{s}, which may be sparse or dense. 
It can also carry the inverse or transpose used for lifting through \texttt{s\_inv}, the original graph membership through \texttt{batch}, and an input-node mask through \texttt{in\_mask} when dense padded inputs are used. 
For dense assignments, \texttt{out\_mask} is derived automatically and marks which pooled supernodes are valid. 
Poolers may attach extra operator-specific fields as needed, which keeps the interface extensible without giving up a common base type.
Table~\ref{tab:select} summarizes the shared fields exposed by this data structure.

\input{tables/select_output}


\paragraph{\texttt{PoolingOutput}} is the universal return type of a pooling layer. It bundles the pooled node features, the pooled connectivity, optional edge weights, the pooled batch vector, the original \texttt{SelectOutput}, and an optional loss dictionary. Its \texttt{mask} property is derived from \texttt{so.out\_mask}. This is the object consumed by downstream message-passing layers, graph-level readout, and lifting. Helper methods such as \texttt{has\_loss}, \texttt{get\_loss\_value()}, and \texttt{as\_data()} make it easier to integrate pooling layers into ordinary PyG code.
Table~\ref{tab:pool} lists the standardized fields consumed by downstream code.

\input{tables/pooling_output}

\section{A taxonomy of graph pooling methods}

We categorize graph pooling methods along two primary, orthogonal axes that highlight key implementation challenges: the presence of learnable parameters (\textbf{trainable vs. non-trainable}) and the structure of the selection matrix (\textbf{sparse vs. dense}). Figure~\ref{fig:dense_sparse_gallery} illustrates these differences through concrete examples of sparse and dense, trainable and non-trainable pooling operators applied to the same input graph.

In the literature, these distinctions are often treated primarily as theoretical properties. In \texttt{tgp}, they translate directly into explicit software workflows. The following subsections explore both axes, directly pairing their theoretical definitions with the practical implementations they demand---specifically, the \texttt{PreCoarsening} pipelines uniquely enabled by non-trainable methods, and the diverse execution modes engineered to accelerate dense assignment operators.

\newcolumntype{M}[1]{>{\centering\arraybackslash}m{#1}}
\begin{figure}[htbp]
\centering
\setlength{\tabcolsep}{3pt}
\begin{tabular}{@{}M{0.2\linewidth} M{0.18\linewidth} M{0.18\linewidth} M{0.18\linewidth} M{0.18\linewidth}@{}}
  \includegraphics[width=\linewidth]{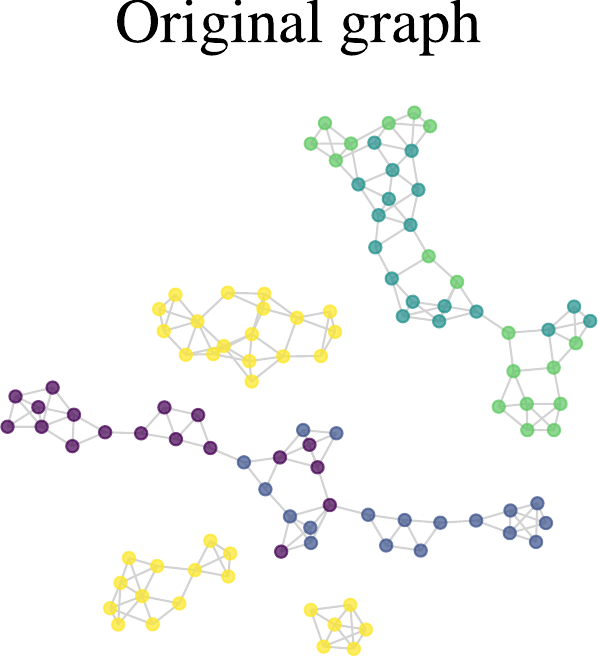}
  & \includegraphics[width=\linewidth]{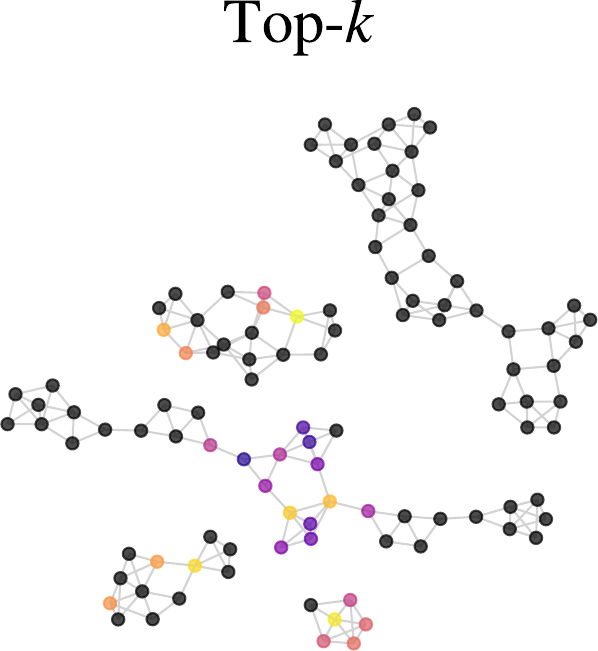}
  & \includegraphics[width=\linewidth]{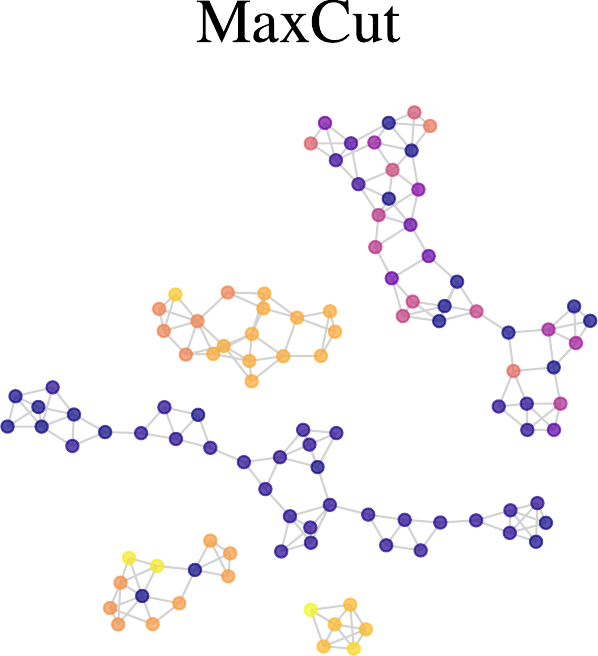}
  & \includegraphics[width=\linewidth]{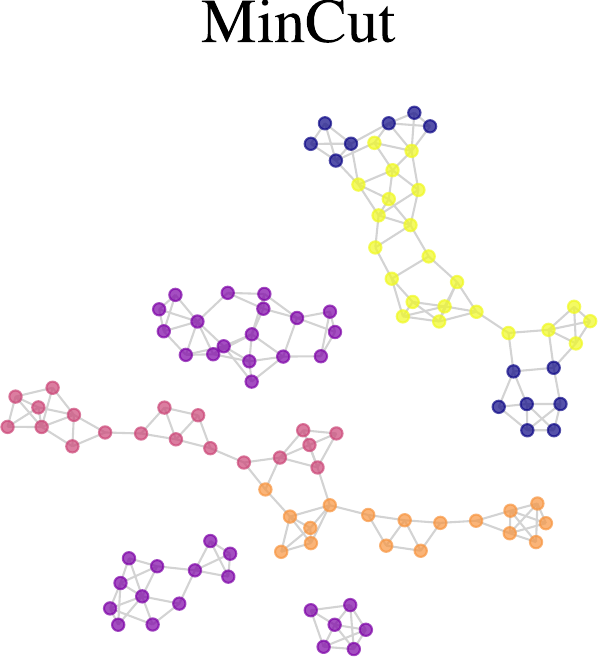}
  & \includegraphics[width=\linewidth]{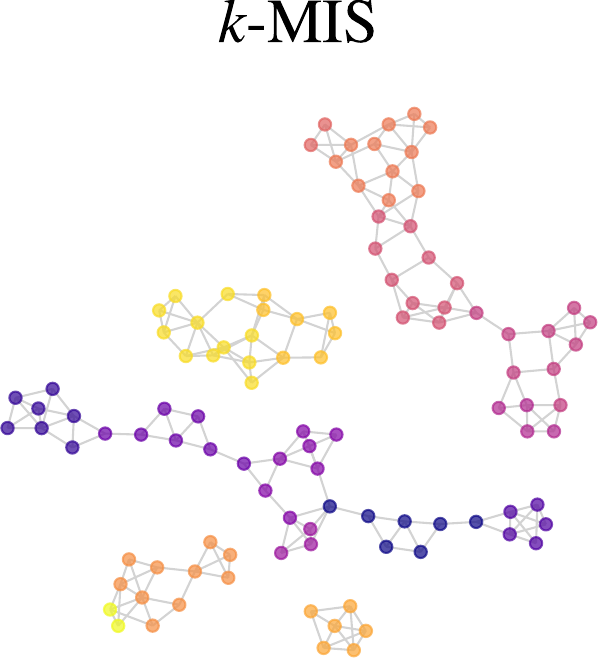} \\
  Selection matrix $\mathbf{S}$:
  & \includegraphics[width=0.6\linewidth]{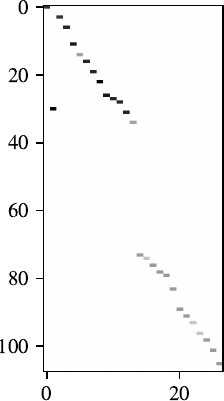}
  & \includegraphics[width=0.6\linewidth]{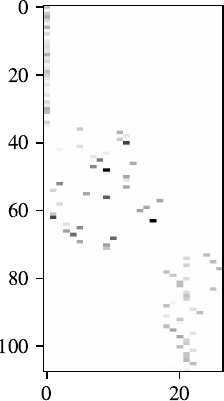}
  & \includegraphics[width=0.6\linewidth]{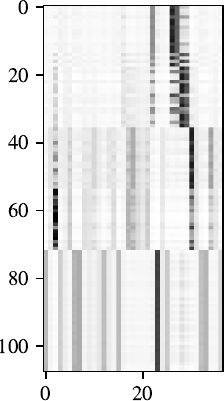}
  & \includegraphics[width=0.6\linewidth]{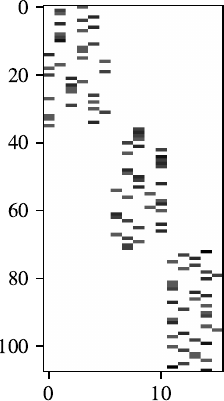} \\
  Coarsened graph $\mathbf{A}'$:
  & \includegraphics[width=\linewidth]{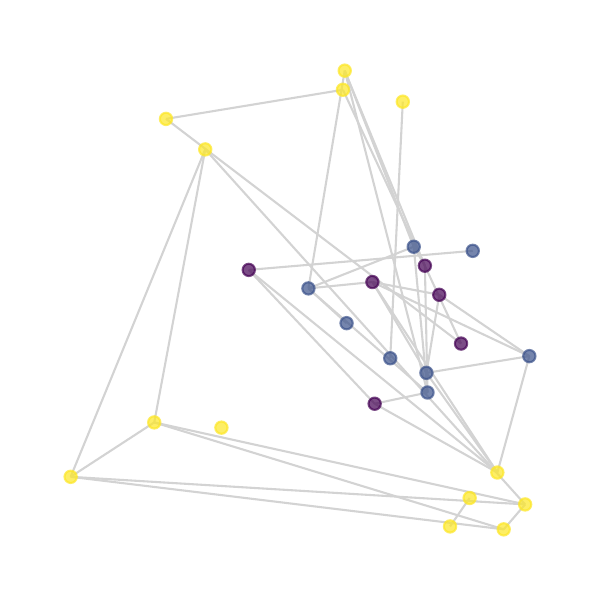}
  & \includegraphics[width=\linewidth]{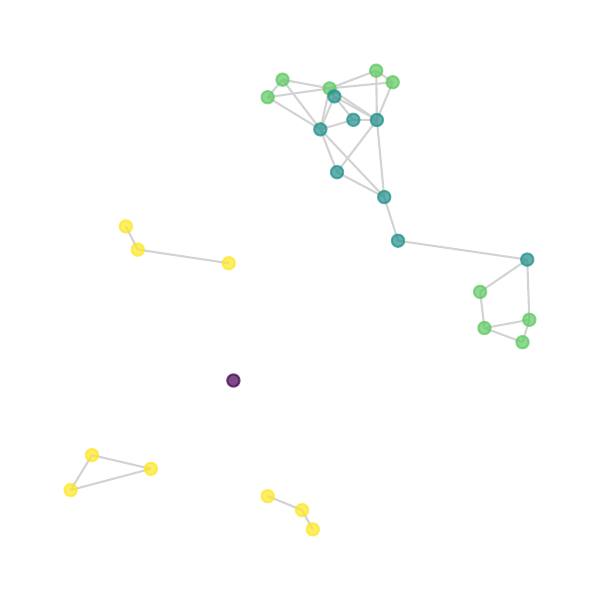}
  & \includegraphics[width=\linewidth]{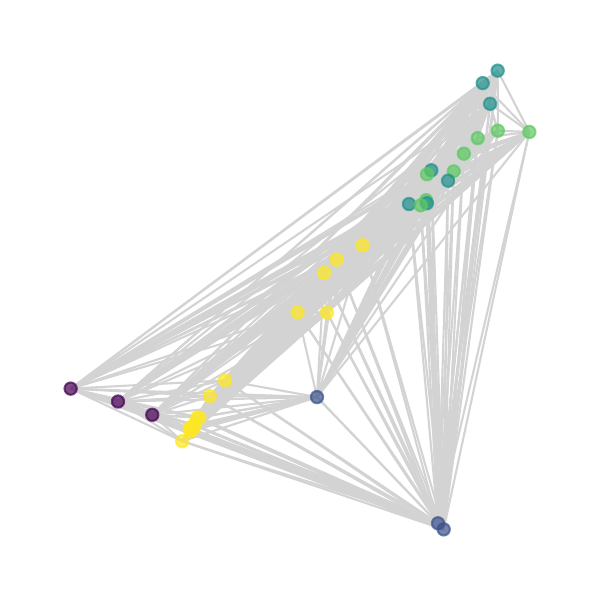}
  & \includegraphics[width=\linewidth]{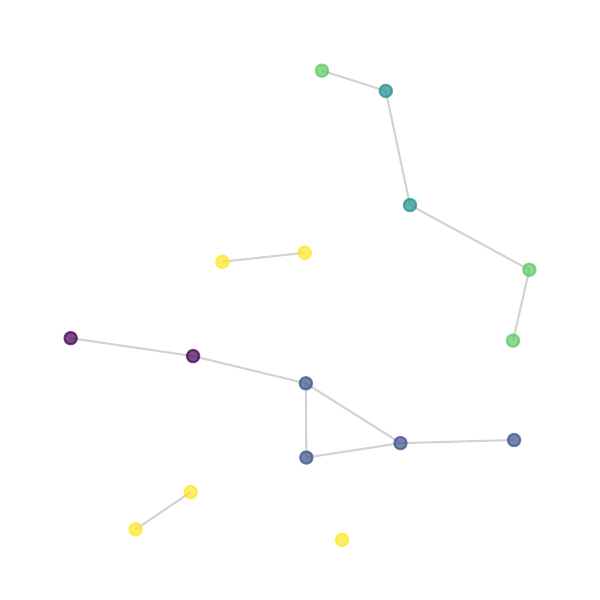} \\
\end{tabular}
\caption{Examples of pooling operators with different assignment structures. The top row shows the original graph and node-to-supernode assignments. The middle row shows the corresponding assignment matrices. The bottom row shows the pooled graphs.}
\label{fig:dense_sparse_gallery}
\end{figure}

\subsection{Sparse vs. dense assignments and execution paths}
\label{app:dense-sparse-pooling}
\paragraph{Sparse vs. dense assignment}
Pooling operators can be further divided based on the structure of the assignment matrix $\mathbf{S}$ they produce. 
This distinction significantly affects memory footprint, computational complexity, and the structure of the pooled graph.
In \textit{Sparse poolers}, the assignment matrix $\mathbf{S} \in \{0,1\}^{N \times K}$ has a number of non-zero entries proportional either to the number of supernodes ($K$) or to the number of nodes in the original graph ($N$).
Methods such as \gls{topk}~\citep{gao2019graph, knyazev2019understanding} learn a scoring function to determine which nodes to select. 
This approach is highly efficient but leads to information loss, as non-selected nodes are discarded entirely.
\textit{Dense poolers}, often called \textit{soft-clustering} poolers, produce a dense assignment matrix $\mathbf{S}$, which gives the $N$ original nodes a ``soft'' membership to each of the $K$ supernodes (clusters). 
Representatives like \gls{diff}~\citep{ying2018hierarchical} and \gls{mincut}~\citep{bianchi2020spectral} learn this differentiable node partitioning guided by auxiliary losses. 
While highly expressive \citep{bianchi2023expr}, these poolers are computationally demanding and typically require specifying a fixed number of clusters $K$, which is problematic for datasets with graphs of variable sizes.

\paragraph{Dense execution modes and sparse-output interoperability}
\label{app:dense-execution-modes}

Dense poolers support two execution modes. When \texttt{batched=True}, a dense pooler accepts sparse graph batches and internally converts them to padded tensors of shape $[B, N_{\max}, F]$ and $[B, N_{\max}, N_{\max}]$, together with an input validity mask of shape $[B, N_{\max}]$. This path matches dense message-passing layers.

When \texttt{batched=False}, supported dense poolers intentionally follow sparse-pooler API semantics: they consume sparse graph inputs (\texttt{edge\_index}, optional \texttt{edge\_weight}, and \texttt{batch}) rather than padded dense tensors. The assignment remains dense, but connectivity stays sparse. It avoids padding overhead and supports variable-size graph batches without constructing a large padded adjacency tensor first. Poolers that support both paths are listed in Appendix~\ref{app:poolers}. \gls{eigen} and \gls{nmf} are implemented as dense poolers but expose only the unbatched path.

\input{listings/batched-vs-unbatched}

Output format is independent from execution mode. If \texttt{sparse\_output=False}, poolers return batched dense pooled graphs. If \texttt{sparse\_output=True}, the pooled result is converted to a block-diagonal sparse representation so downstream layers can continue with sparse \gls{pyg} message passing. This enables mixed architectures in which the selector is dense but post-pooling message passing remains sparse.

The four combinations of \texttt{batched} and \texttt{sparse\_output} are the main entry points. Listing~\ref{lst:dense-mode-configs} shows a compact inspection loop for this interface and the corresponding downstream branch.

\subsection{Trainable vs. non-trainable methods and pre-coarsening}
\paragraph{Trainable vs. non-trainable }
A fundamental distinction among pooling operators is whether the coarsened graph is obtained either through trainable or pre-computed operations. 
\textit{Non-trainable methods} usually rely on graph-theoretic algorithms that account only for the graph's topology. 
This implies that the pooled graph's topology does not depend on the node features and the downstream task. 
Examples include methods based on spectral clustering~\citep{dhillon2007weighted}, largest eigenvector vertex selection~\citep{bianchi2020hierarchical}, and maximal independent sets~\citep{bacciu2023pooling}. 
Since the coarsening is feature-independent, the pooled graphs can be pre-computed, simplifying training at the cost of expressivity, as the pooling strategy does not adapt to the task at hand. 
\textit{Trainable methods}, instead, incorporate learnable parameters that are optimized end-to-end with the rest of the model. 
A trainable pooler can identify the most important nodes or substructures for the task at hand, often leading to superior performance at the cost of computational complexity.

\paragraph{Offline Hierarchy with Pre-coarsening}
\label{app:workflow-offline}

\gls{tgp} provides \texttt{PreCoarsening}, a dataset transform that computes a hierarchy of pooled graphs before training and attaches the result to each \gls{pyg} \texttt{Data} object. 
\input{listings/precoarsening}
It accepts a single pooler, a repeated sequence of poolers, or per-level configurations. The batched side of this workflow is handled by \texttt{PoolDataLoader} and \texttt{PooledBatch}, which collate the original graphs together with the stored coarsened levels. 
The same interface therefore covers ordinary repeated rollouts and operators whose natural output is a sequence of coarsened graphs.
Listing~\ref{lst:precoarsening-schedules} shows three schedules built with the same interface: repeated use of one pooler, the same family with different parameters at each level, and a mixed hierarchy that chains different operators inside one pre-coarsening rollout.

The multi-level design matters for operators such as \gls{sep} because they produce a hierarchy as part of the operator itself, not as the result of repeating a one-level pooler. In that case, pre-coarsening exposes the intermediate graphs directly and keeps the hierarchy in the public API. The same rollout also covers ordinary deterministic schedules, such as a two-level \gls{ndp} hierarchy or a mixed schedule that changes poolers from level to level. These workflows belong to the public API rather than to one-off preprocessing code.

\section{Benchmark Workflows and Architectures}
\label{app:workflow-listings}

The empirical sections of this work rely on three benchmark families that all use the same pooler API---each forward pass returns a \texttt{PoolingOutput} with coarsened tensors, batching metadata, and optional auxiliary losses---but differ in how those outputs are consumed.
For \emph{graph-level} prediction, pooled node features feed further message passing and then a global readout.
For \emph{node classification}, the model pools to a bottleneck, processes the coarse graph, and \emph{lifts} back to the original nodes before predicting labels.
For \emph{unsupervised vertex clustering}, a dense trainable pooler is trained only on its auxiliary objectives; hard partitions are read from the assignment matrix~$\mS$.
The subsections below give compact reference implementations of these public workflows. The concrete experimental configurations are stated with the corresponding results tables in Appendix~\ref{app:evaluation-overview}.

\subsection{Graph-level readout}
\label{app:workflow-graph}

\begin{figure}[!ht]
    \centering
    \includegraphics[width=0.6\linewidth]{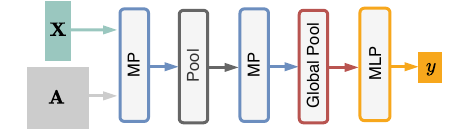}
    \caption{Architecture for graph-level tasks.}
    \label{fig:graph-level-diag}
\end{figure}

Graph-level models interleave message passing with pooling and end with a
readout that maps pooled node features to a graph embedding
(Figure~\ref{fig:graph-level-diag}).

Listing~\ref{lst:graph-level-workflow} shows the forward pass used by the
benchmark: a pooler from \texttt{get\_pooler} between two \gls{gin}
\citep{xu2018powerful} message-passing blocks, branching on
\texttt{is\_dense} and \texttt{sparse\_output} to pick sparse or dense
convolutions after pooling, and a sum \texttt{GlobalReduce} readout followed
by a \gls{mlp} head.
Auxiliary losses, when present in \texttt{PoolingOutput}, are added to the
task loss so trainable poolers optimize their own objectives jointly with
supervision.

\input{listings/graph-level}

\subsection{Node classification with pooling and lift}
\label{app:workflow-node}

Node-level benchmarks follow an encoder--pool--bottleneck--lift--decoder pattern (as in Graph U-Net--style hierarchies~\citep{gao2019graph}): the pooler returns \texttt{PoolingOutput} and the metadata needed to lift coarse features back to the original nodes before per-node prediction.
\begin{figure}[!ht]
    \centering
    \includegraphics[width=0.75\linewidth]{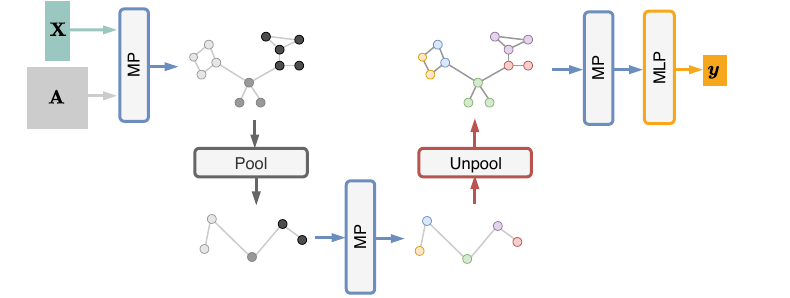}
    \caption{Node classification architecture.}
    \label{fig:node-class-diag}
\end{figure}

Listing~\ref{lst:node-class-pool-lift} condenses the node-level workflow: a \texttt{GINConv} encoder, dense \texttt{mincut} from \texttt{get\_pooler}, a \texttt{DenseGINConv} bottleneck on \texttt{PoolingOutput} with \texttt{out.mask}, then a \texttt{lifting=True} pass driven by \texttt{out.so} and the node/supernode batch vectors before the final \texttt{DenseGINConv} head.
Training uses masked NLL on \texttt{train\_mask} and adds pooling auxiliaries with \texttt{sum(out.get\_loss\_value())} when present.

\input{listings/node-level}

\subsection{Unsupervised vertex clustering}
\label{app:workflow-clustering}

Clustering benchmarks stack a shallow message-passing encoder in front of a trainable dense pooler.
\begin{figure}[!ht]
    \centering
    \includegraphics[width=0.5\linewidth]{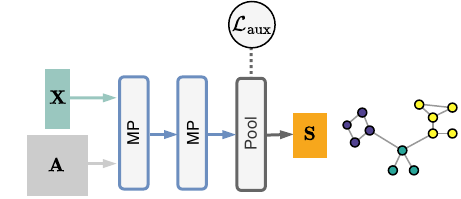}
    \caption{Node clustering architecture.}
    \label{fig:clust-diag}
\end{figure}
Optimization uses only the auxiliary objectives packaged in \texttt{PoolingOutput}; the assignment matrix~$\mS$ defines soft memberships during training and yields hard partitions at evaluation time.

Listing~\ref{lst:clustering-unsupervised} sketches the training forward pass: stacked \texttt{ARMAConv} layers, a dense pooler from \texttt{get\_pooler}, and the auxiliary loss sum.
The control flow matches the library example \texttt{clustering.py}.

\input{listings/clustering}

\section{\texttt{AggrReduce} and \texttt{GlobalReduce}}
\label{app:aggr-reduce}

Readout and the \texttt{Reduce} stage of a pooler perform the same operation at
two different scales.
A \texttt{Reduce} stage consumes an assignment $\mS$ and collapses each group of
original nodes into one supernode feature vector.
A graph-level readout is the same mapping with a single group per graph: all
nodes of a graph are collapsed into one embedding.
\gls{tgp} exposes this symmetry through two modules that share the same
aggregation backend.

\paragraph{\texttt{AggrReduce}}
wraps any \gls{pyg} \texttt{Aggregation} and plugs it into the \texttt{Reduce}
slot of an \texttt{SRCPooling} operator.
The wrapped aggregator receives the node features and the group indices derived
from the \texttt{SelectOutput}, and returns one feature row per supernode.
Because the aggregation object is a standard \gls{pyg} module, the full family
of aggregators is available: \texttt{sum}, \texttt{mean}, \texttt{max},
\texttt{std}, \texttt{softmax}, \texttt{powermean}, \texttt{lstm},
\texttt{set2set}, \texttt{set\_transformer}, \texttt{graph\_multiset\_transformer},
among others.
Swapping the reducer therefore changes how supernode features are computed
without touching the \texttt{Select} or \texttt{Connect} stages of the pooler,
and without any change to the surrounding model code.

\paragraph{\texttt{GlobalReduce}}
is the corresponding all-to-one module used at graph-level readout.
It accepts either a \texttt{batch} vector (sparse batching) or a \texttt{mask}
(dense padded batching), which makes it compatible with both branches of the
graph-level workflow in Appendix~\ref{app:workflow-graph}.
Its \texttt{reduce\_op} argument selects the same family of aggregators as
\texttt{AggrReduce}, so a model can use the same aggregation primitive inside
the pooler and at readout.

Listing~\ref{lst:aggr-reduce} replaces the default reducer of a sparse
\gls{topk} pooler with an \texttt{AggrReduce} wrapping \texttt{set2set}, and
instantiates a \texttt{GlobalReduce} that applies the same \texttt{set2set}
aggregation at graph-level readout.
Sharing the aggregation primitive between the two stages keeps tensor contracts
aligned across the architecture, so a heavier readout can be tried inside the
pooler too with no further change to the model.
The rest of the forward pass in Listing~\ref{lst:graph-level-workflow} is left
unchanged: the pooler still returns a \texttt{PoolingOutput}, the post-pooling
convolution still consumes \texttt{out.x} and \texttt{out.edge\_index}, and the
readout still produces one embedding per graph.

\input{listings/aggr-reduce}

\section{Modularity and extensibility}
Each pooler exposes boolean flags that let model code select the correct execution path without inspecting operator internals. Table~\ref{tab:flags-workflow} lists the three main flags and their typical use in a training loop.

\input{tables/flags}

\label{app:workflow-custom}
The \gls{srcl} framework's decomposition of pooling into distinct stages naturally provides a modular software design where the \texttt{Select}, \texttt{Reduce}, \texttt{Connect}, and \texttt{Lift} stages are implemented as interchangeable components. 
This architecture is the backbone of \gls{tgp}'s extensibility. 
To make this concrete, in Listing~\ref{lst:topk_implementation} we sketch the implementation of the \gls{topk} operator, which is constructed by combining specific \gls{srcl} components in its \texttt{\_\_init\_\_} method. 
The \texttt{forward} pass then orchestrates the whole pooling logic through a sequence of calls to these components.
It is easy to observe that, for example, to experiment with a new node scoring mechanism, one could simply write a new \texttt{Select} module and plug it into the \texttt{TopkPooling} class, reusing the existing components that perform \texttt{Reduce} and \texttt{Connect} operations.

Similarly, one could instantiate hybrid poolers by replacing the standard sparse connector with a \texttt{KronConnect} module. Listing~\ref{lst:custom_layer} illustrates how to swap the connector of a \gls{topk} pooler with a single line of code.

\input{listings/topk}

\input{listings/swap}

\section{Implemented Pooling Layers}
\label{app:poolers}
\label{app:pooler-cheatsheet}

\acrlong{tgp} covers pooling operators across all four regions of the sparse/dense and trainable/deterministic taxonomy:
\begin{itemize}
\item \textbf{Sparse trainable methods:} \gls{topk} \citep{gao2019graph,knyazev2019understanding}, \gls{sag} \citep{lee2019self}, \gls{asap} \citep{ranjan2020asap}, \gls{pan} \citep{ma2020path}, \gls{ecpool} \citep{diehl2019edge,landolfi2022revisiting}, and \gls{maxcut} \citep{abate2025maxcutpool}.
\item \textbf{Sparse deterministic methods:} \gls{graclus} \citep{dhillon2007weighted}, \gls{ndp} \citep{bianchi2020hierarchical}, \gls{kmis} \citep{bacciu2023pooling}, and \gls{sep} \citep{wu2022structural}.
\item \textbf{Dense trainable methods:} \gls{diff} \citep{ying2018hierarchical}, \gls{mincut} \citep{bianchi2020spectral}, \gls{dmon} \citep{tsitsulin2020graph}, \gls{jbgnn} \citep{bianchi2022simplifying}, \gls{hosc} \citep{duval2022higher}, \gls{bnpool} \citep{castellana2025bnpool}, \gls{acc} \citep{hansen2023total}, and \gls{lapool} \citep{noutahi2019towards}.
\item \textbf{Dense deterministic methods:} \gls{nmf} \citep{bacciu2019non} and \gls{eigen} \citep{ma2019eigenpool}.
\end{itemize}

The appendix cheatsheet in Table~\ref{tab:pooler_features} consolidates this inventory and reports the user-facing traits that affect model integration: assignment type, trainability, auxiliary pooling loss, dense execution support, and pre-coarsening eligibility.

\input{tables/poolers}

\section{Dataset Details}
\label{app:dataset-details}

\subsection{Node Clustering Datasets}
\label{app:node_clust_data}

For the unsupervised clustering task, we use four well-known citation network benchmarks~\citep{yang2016revisiting, fu2020magnn} and a synthetic dataset designed to provide a controlled evaluation environment.
The citation network datasets are loaded using the API provided by \gls{pyg}.
The synthetic Community dataset is generated with PyGSP~\citep{defferrard2017pygsp} and consists of a graph sampled from a stochastic block model with 5 communities.
The statistics for all datasets are reported in Table~\ref{tab:clustering_datasets}.

\input{tables/datasets/clustering}

\subsection{Node Classification Datasets}
\label{app:node_class_data}

The node classification experiments are conducted on the five large-scale heterophilic graphs introduced by~\citep{platonov2023a}. These datasets are loaded using the API provided by \gls{pyg} and come with 10 predefined public splits for training, validation, and testing. The statistics of the five datasets are reported in Table~\ref{tab:node_class_datasets_app}. The column $h(\mathcal{G})$ reports the class-insensitive edge homophily ratio~\citep{lim2021large}, where lower values indicate a higher degree of heterophily.

\input{tables/datasets/node-level}

\subsection{Graph-Level Task Datasets}
\label{app:graph_task_data}

For the benchmark on graph-level tasks, we consider graph classification datasets from diverse domains and with different graph characteristics. 
In particular, we included in our evaluation bioinformatics datasets from the TUDatasets collection~\citep{morris2020tudataset} (NCI1), social networks (Reddit-binary), and synthetic benchmarks designed to test specific model capabilities (expwl1~\citep{bianchi2023expr}, Multipartite~\citep{abate2025maxcutpool}, \gls{gcb}~\citep{bianchi2022pyramidal}, and CSBM~\citep{deshpande2018contextual}). 
The statistics for each dataset are provided in Table~\ref{tab:graph_class_datasets_app}.

\input{tables/datasets/graph-level}

\section{Complete Experimental Results}
\label{app:additional_results}
\label{app:evaluation-overview}

The experiments below were all run on the same software stack: PyTorch~\textsc{2.4.1} with PyG~\textsc{2.6.1} on CUDA~\textsc{12.1}, together with \texttt{tgp}~\textsc{1.0.2}.

\subsection{Node Clustering Results}

This experiment evaluates the ability of dense pooling operators to identify, within a single graph and in a purely unsupervised fashion, communities that align well with the node labels.
We train a \gls{gnn} to generate a node partition by optimizing only the auxiliary losses provided by assessed pooling layers, without any supervised signal.
The partitions are given by the assignment matrix $\mS$ of six dense pooling operators: \gls{acc}~\citep{hansen2023total}, \gls{diff}, \gls{dmon}~\citep{tsitsulin2020graph}, \gls{hosc}~\citep{duval2022higher}, \gls{jbgnn}~\citep{bianchi2022simplifying}, and \gls{mincut}.
The schematic and reference listing appear in Appendix~\ref{app:workflow-clustering}; datasets are summarized in Appendix~\ref{app:node_clust_data}.

\phantomsection\label{app:setup_clustering}
The benchmark fixes the architecture illustrated in Figure~\ref{fig:clust-diag} and Listing~\ref{lst:clustering-unsupervised}: two \gls{mp} layers implemented as \texttt{ARMAConv}~\citep{bianchi2021graph} with ELU activations, mapping inputs to a 32-dimensional embedding before the dense pooling layer.
Training optimizes only the auxiliary losses returned in \texttt{PoolingOutput}.
We train for up to 2000 epochs using the Adam optimizer~\citep{kingma2014adam} with a learning rate of $5 \cdot 10^{-4}$ and the ReduceLROnPlateau scheduler\footnote{\href{https://docs.pytorch.org/docs/stable/generated/torch.optim.lr_scheduler.ReduceLROnPlateau.html}{https://docs.pytorch.org/docs/stable/generated/torch.optim.lr\_scheduler.ReduceLROnPlateau.html}}, with early stopping (patience of 500 epochs).
After training, we extract hard cluster assignments via an $\operatorname{argmax}$ operation on the matrix $\mS$ and evaluate quality using the Normalized Mutual Information (NMI).

\input{tables/results/clustering}

We note that changing the hyperparameters in an unsupervised task like this one can significantly modify the outcome.
However, tuning the hyperparameters in a clustering setting without relying on supervised information is not straightforward and out of scope for this evaluation.

Table~\ref{tab:clustering_results} shows that operators with strong graph-theoretic inductive biases, such as \gls{acc} and \gls{dmon}, excel at creating node partitions that align with the true class of the nodes.
Optimizing well-established graph partitioning objectives, such as the minimum cut and the graph modularity, works particularly well in homophilic graphs where the nodes of a community have similar class labels.
In contrast, \gls{diff} consistently underperforms in terms of \gls{nmi} as it optimizes losses inspired by heuristics rather than graph-theoretical objectives: a link prediction loss that encourages connected nodes to be in the same cluster and an entropy term that prevents cluster assignments from being too smooth. 
Without the primary learning signal provided by a supervised objective, these regularizers alone are often insufficient to guide the model toward learning meaningful partitions. 
In summary, the results underscore that in unsupervised tasks, the result depends heavily on the alignment between the objectives optimized by its auxiliary losses, the structure of the graph, and the properties of its nodes.

\subsection{Node Classification Results}

This experiment evaluates the performance of the pooling operators in transductive node classification tasks on predominantly heterophilic graphs.
The schematic and reference listing appear in Appendix~\ref{app:workflow-node}; datasets are summarized in Appendix~\ref{app:node_class_data}.

\phantomsection\label{app:setup_node_class}
We use the hierarchical layout in Figure~\ref{fig:node-class-diag};
Listing~\ref{lst:node-class-pool-lift} is a compact MinCut example that highlights the lifting control flow.
The benchmark instantiates this layout as  \gls{mp}-Pool-\gls{mp}-Unpool-\gls{mp}-Readout with \gls{gin}~\citep{xu2018powerful} (32 hidden units, ReLU) and an \gls{mlp} readout with one hidden layer and dropout rate $0.1$, following Graph U-Net--style encoder--decoder designs~\citep{gao2019graph}.
The model is trained for up to 20,000 epochs by jointly minimizing the cross-entropy loss and any auxiliary loss, using the Adam optimizer with a learning rate of $5 \cdot 10^{-4}$, a scheduler, and early stopping (patience of 2,000 epochs).
For each dataset, we report the performance in terms of accuracy (Roman-Empire, Amazon-Ratings) and AUROC (Minesweeper, Tolokers, Questions) over 10 public data folds, following the same setting proposed in the original paper~\citep{platonov2023a}.

\input{tables/results/node-level}

The results in Table~\ref{tab:node_classification_results} demonstrate that \gls{kmis} and \gls{maxcut} achieve the strongest overall performance across the evaluated datasets.
Both methods succeed due to an inductive bias that favors a spatially uniform subsampling of the graph, explicitly identifying supernodes that remain largely disconnected from each other. 
In a heterophilic setting, where structurally adjacent nodes frequently hold different class labels, drawing independent supernodes across the graph serves to prevent the premature aggregation of dissimilar neighborhoods.
By maintaining the structural diversity of the original graph during the pooling phase (see Figure \ref{fig:dense_sparse_gallery}), these operators produce a high-quality bottleneck representation that can be effectively un-pooled via the \texttt{Lift} operation for accurate node-level predictions.

\subsection{Graph-Level Classification Results}
\label{app:setup_graph_class}
The schematic and reference listing appear in Appendix~\ref{app:workflow-graph}; datasets are summarized in Appendix~\ref{app:graph_task_data}.

In the experiments we use a hierarchical \gls{gin} stack (32 hidden channels, ELU) before and after pooling, switching the post-pooling block to \texttt{DenseGINConv} when the pooler is dense, then global sum readout and a 3-layer \gls{mlp} with dropout $0.5$.
The same \texttt{[Pool - MP]} block can be repeated in deeper architectures.
For graph classification, we optimize either cross-entropy or binary cross-entropy, depending on the dataset.
If a pooler provides auxiliary losses, they are added to the task loss.
We train for up to 1000 epochs with the Adam optimizer (learning rate $1 \cdot 10^{-4}$), a scheduler, and early stopping (patience 300).
For datasets with public splits, we perform 10 runs; otherwise, we use 10-fold cross-validation.
We report test classification accuracy.

\input{tables/results/graph-level}

The aggregate ``Score'' column in Table~\ref{tab:graph_classification_results} counts how often each operator lands at or near the front of the pack on these six benchmarks; it is a coarse summary, but it makes a point visible at a glance:
the contrast with vertex clustering (Table~\ref{tab:clustering_results}) is sharp.
\gls{acc} and \gls{dmon}, which lead on citation-style clustering under auxiliary losses alone, never claim a column-wise best here and sit in the mid-field: objectives that organize unlabeled communities do not transfer into a supervised graph-classification head without re-tuning, and may fight the \gls{gin} readout used in this protocol.

Dense trainable poolers (\gls{mincut}, \gls{jbgnn}, \gls{hosc}, \gls{bnpool}, \gls{lapool}, \gls{diff}) cluster toward respectable averages across molecular and social graphs, whereas sparse trainable and deterministic poolers swing harder.
\gls{kmis} and \gls{maxcut} dominate the synthetic WL and block-model style columns where aggressive, well-spread subsampling matches the inductive bias of the generator; \gls{sep} and \gls{ecpool} peak on GCB-H and CSBM respectively when the graph family aligns with their construction.
That pattern supports a practical rule: soft-assignment dense layers are a robust default for broad screening, while sparse or structure-driven poolers reward explicit alignment between the operator and the dataset generator or domain.

\subsection{Efficiency Results}
\phantomsection\label{app:setup_efficiency}

We measure the per-batch wall-clock cost of pooling in two settings that mirror the main benchmarks: \emph{caching} on node classification and \emph{pre-coarsening} on graph classification. We keep the architectures of the main benchmarks, and sweep the non-trainable poolers on Amazon-ratings, Roman-empire, and Tolokers (node) and on MUTAG, Reddit-binary, and GCB-Hard (graph).

A callback wraps each pooling layer with forward and full-backward hooks timed via \texttt{time.perf\_counter()}; per epoch we log the mean forward and mean backward duration at the first pooler and plot their sum, averaged across 100 batches. In both node and graph level settings the first 2 epochs are warmup and discarded. All timings are collected on a single \textsc{NVIDIA RTX 6000 Ada Generation} with an
\textsc{AMD Ryzen 9 5900X 12-Core Processor} and \textsc{32}~GB of RAM.

\begin{figure}[htbp]
\centering
\includegraphics[width=\linewidth]{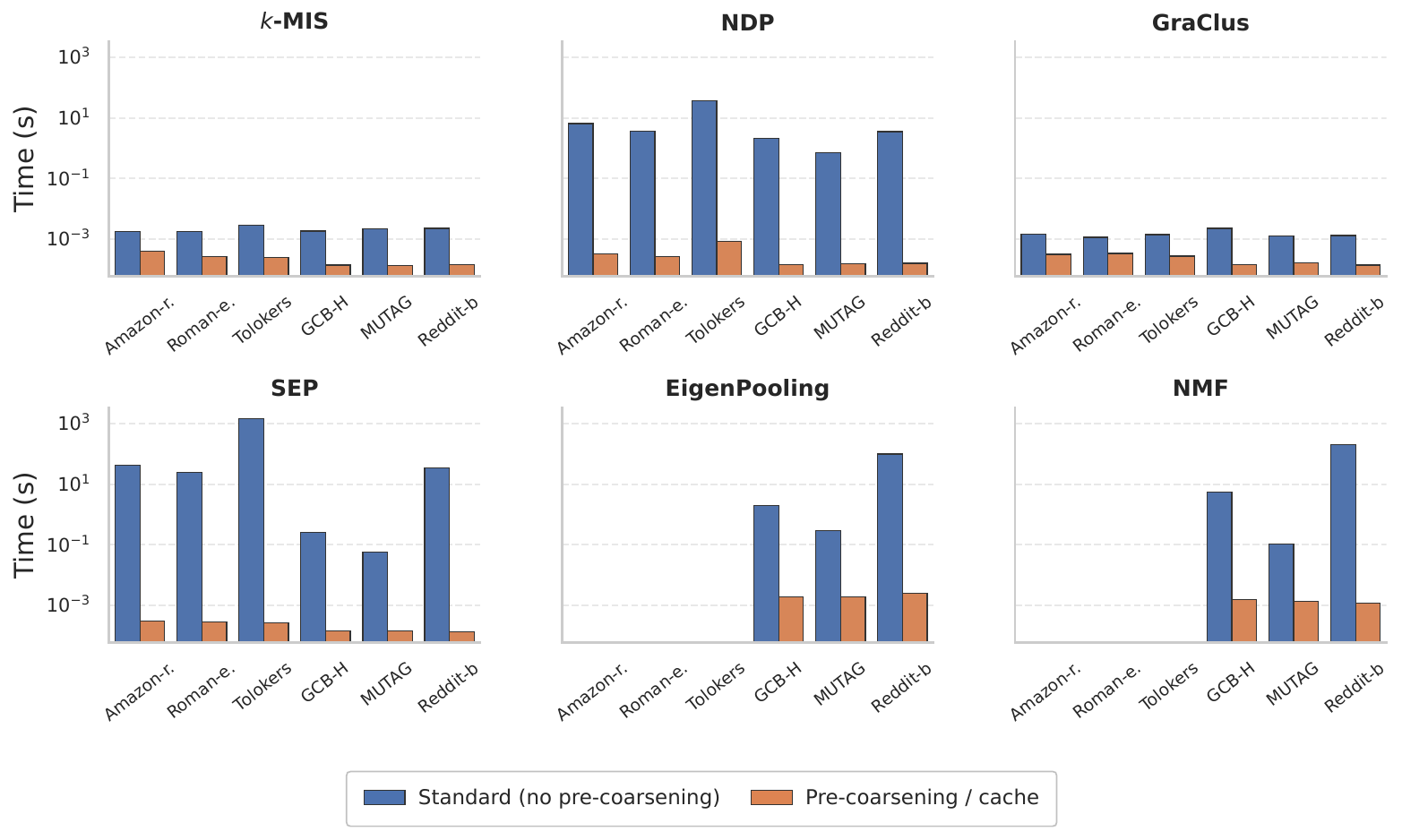}
\caption{Bar chart of mean batch forward and backward pass time for caching (Amazon Ratings, Roman Empire, Tolokers) and for dataset pre-coarsening (GCB-H, MUTAG, Reddit binary); paired bars compare the online pooler path with caching or pre-coarsening. Missing bars mean no reliable timing was available (no stable converged run). This mainly affects \gls{eigen} and \gls{nmf} on the large node classification graphs: their \texttt{Select} paths operate from a dense adjacency view and invoke spectral clustering or nonnegative matrix factorization, which becomes costly and numerically delicate as the number of nodes grows, so timings are omitted where runs did not complete cleanly.}
\label{fig:efficiency_results}
\end{figure}

Results are shown in Figure~\ref{fig:efficiency_results}.

\subsection{Dense-Mode Execution Paths}
\label{app:dense-modes}

\label{app:setup_dense_modes}

We compare batched padded dense execution against the unbatched sparse-connectivity path for dense poolers under matched configurations, recording peak GPU memory and average epoch runtime on two benchmark regimes:

\paragraph{Grid graphs (grid side axis).}
We use 2D grid graphs (PyGSP \texttt{Grid2d}): square $N \times N$ layouts with four-neighbor connectivity.
We increase $N$ over $\{50, 75, 100, 125, 150\}$, so the number of nodes grows quadratically while the topology remains regular, isolating scaling of runtime and peak memory for the two dense-pooler execution modes.

\paragraph{Synthetic multi-graph classification (batch size axis).}
We benchmark on a synthetic multi-graph binary classification dataset.
Each graph is a connected Erd\H{o}s--R\'enyi graph (PyGSP \texttt{ErdosRenyi}) with between $10$ and $500$ nodes, edge probability $p = 0.02$, undirected edges, and no self-loops.
Training uses mini-batches of variable-sized graphs; we vary the batch size to stress padded, batched dense tensor layouts against the unbatched sparse-connectivity path.

Each (pooler, mode) pair is trained using the same architectures as the main benchmarks: a clustering head on Grid2D and the hierarchical graph classifier on the multi-graph dataset. We sweep the dense poolers, each instantiated once with \texttt{(batched=True, sparse\_output=False)} and once with \texttt{(batched=False, sparse\_output=True)}; all other hyperparameters are shared across modes. Runtime is the same per-batch mean of forward plus backward time at the first pooler as in Appendix~\ref{app:setup_efficiency}. Peak GPU memory is recorded by a callback that calls \texttt{torch.cuda.synchronize} and \texttt{reset\_peak\_memory\_stats} before each pooler forward and reads \texttt{max\_memory\_allocated} after synchronizing again, so the reported figure is the pooler-local peak. The first $2$ epochs are warmup and discarded. All measurements are taken on the same hardware and software stack as in Appendix~\ref{app:setup_efficiency}.

\begin{figure}[htbp]
\centering
\includegraphics[width=\linewidth]{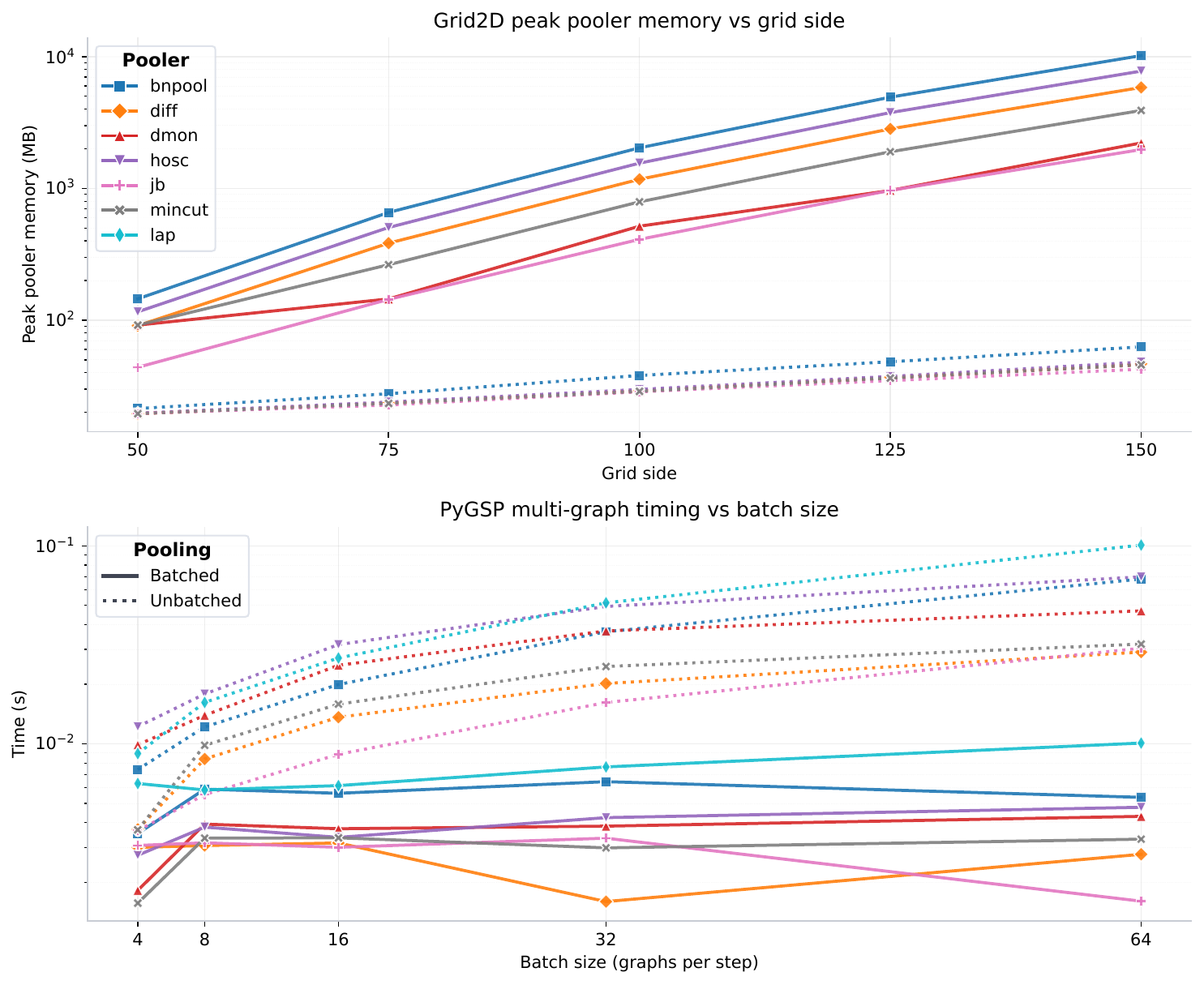}
\caption{%
Batched versus unbatched execution paths for dense poolers. Batched mode
(\texttt{batched=True, sparse\_output=False}) pads each batch to a common node
count and runs the pooler on a dense tensor, trading memory for vectorization;
unbatched mode (\texttt{batched=False, sparse\_output=True}) keeps the sparse
batched connectivity and iterates over graphs, avoiding padding overhead.
\textbf{Top:} peak pooler memory on PyGSP Grid2D graphs as the grid side grows,
showing the super-linear memory cost of the batched path.
\textbf{Bottom:} mean per-batch forward-plus-backward pooler time on a
synthetic Erd\H{o}s--R\'enyi multi-graph task as the batch size grows, showing
that batched execution stays essentially flat while unbatched execution grows
with the batch. Solid lines: batched; dotted lines: unbatched. Markers denote
the pooler.}
\label{fig:batched-unbatched-paths}
\end{figure}

Results are reported in Figure~\ref{fig:batched-unbatched-paths}.

%% file: tables/select_output.tex

\begin{table}[htbp]
\centering
\small
\setlength{\tabcolsep}{6pt}
\renewcommand{\arraystretch}{1.1}
\begin{tabularx}{\linewidth}{>{\raggedright\arraybackslash}p{0.30\linewidth}X}
\toprule
\textbf{Field} & \textbf{Description} \\
\midrule
\texttt{s} & Assignment matrix, sparse or dense \\
\texttt{s\_inv} & Cached inverse or transpose of \texttt{s} used for lifting \\
\texttt{batch} & Batch vector for pooled representations \\
\texttt{in\_mask} & Dense-input validity mask for batched dense assignments \\
\texttt{out\_mask} & Output mask marking valid pooled supernodes (for dense assignments) \\
\texttt{\_extra\_args} & Names of additional pooler-specific attributes attached dynamically \\
\bottomrule
\end{tabularx}
\caption{Components of \texttt{SelectOutput}.}
\label{tab:select}
\end{table}

%% file: tables/pooling_output.tex
\begin{table}[htbp]
\centering
\small
\setlength{\tabcolsep}{6pt}
\renewcommand{\arraystretch}{1.1}
\begin{tabularx}{\linewidth}{>{\raggedright\arraybackslash}p{0.30\linewidth}X}
\toprule
\textbf{Field} & \textbf{Description} \\
\midrule
\texttt{x} & Pooled node features \\
\texttt{edge\_index} & Pooled graph connectivity \\
\texttt{edge\_weight} & Edge weights for the pooled graph (if available) \\
\texttt{batch} & Batch vector for mini-batched graphs \\
\texttt{so} & The originating \texttt{SelectOutput} object \\
\texttt{loss} & Loss dictionary from the pooler, if any \\
\texttt{mask} & Derived validity mask from \texttt{so.out\_mask} \\
\bottomrule
\end{tabularx}
\caption{Components of \texttt{PoolingOutput}.}
\label{tab:pool}
\end{table}

%% file: listings/batched-vs-unbatched.tex
\begin{lstlisting}[language=Python, caption={Pseudo-code for the four dense execution and output modes.}, label={lst:dense-mode-configs}, style=tgpcode]
from tgp.poolers import get_pooler
from tgp.reduce import GlobalReduce

configs = [
    ("Batched Dense", dict(batched=True, sparse_output=False)),
    ("Batched Sparse", dict(batched=True, sparse_output=True)),
    ("Unbatched Dense", dict(batched=False, sparse_output=False)),
    ("Unbatched Sparse", dict(batched=False, sparse_output=True)),
]

readout = GlobalReduce(reduce_op="sum")

for name, config in configs:
    pooler = get_pooler(
        "mincut",
        in_channels=hidden_channels,
        k=10,
        **config,
    )
    out = pooler(
        x=batch.x,
        adj=batch.edge_index,
        edge_weight=getattr(batch, "edge_weight", None),
        batch=batch.batch,
    )

    if pooler.sparse_output:
        x_next = sparse_conv(out.x, out.edge_index, out.edge_weight)
        graph_emb = readout(x_next, batch=out.batch)
    else:
        x_next = dense_conv(out.x, out.edge_index, mask=out.mask)
        graph_emb = readout(x_next, mask=out.mask)
\end{lstlisting}

%% file: listings/precoarsening.tex
\begin{lstlisting}[language=Python, caption={Pre-coarsening with repeated, per-level, and mixed pooler schedules.}, label={lst:precoarsening-schedules}, style=tgpcode]
from torch_geometric.datasets import TUDataset
from tgp.data import PreCoarsening, PoolDataLoader
# Same pooler at every level.
sep_levels = PreCoarsening(poolers=["sep", "sep", "sep"])
# Same family, different configuration at each level.
nmf_levels = PreCoarsening(
    poolers=[
        ("nmf", {"k": 8}),
        ("nmf", {"k": 4}),
    ])
# Mixed hierarchy: NDP first, then EigenPooling.
mixed_levels = PreCoarsening(
    poolers=[
        "ndp",
        ("eigen", {"k": 4, "num_modes": 3}),
    ])
dataset = TUDataset(
    root="/tmp/MUTAG_mix",
    name="MUTAG",
    pre_transform=mixed_levels,
)
loader = PoolDataLoader(dataset, batch_size=32, shuffle=True)
\end{lstlisting}

%% file: listings/graph-level.tex







\begin{lstlisting}[language=Python, caption={Graph-level forward pass used by the benchmark.}, label={lst:graph-level-workflow}, style=tgpcode]
from torch_geometric.nn import GINConv, DenseGINConv
from tgp.poolers import get_pooler
from tgp.reduce import GlobalReduce
pooler = get_pooler(pooler_name,
                    in_channels=hidden_channels,
                    **pooler_kwargs)
                    
# Branch the post-pooling block on the pooler's flags.
use_dense_branch = pooler.is_dense and not pooler.sparse_output
conv_pre  = GINConv(in_channels, hidden_channels)
conv_post = (DenseGINConv if use_dense_branch else GCNConv)(
    hidden_channels, hidden_channels)
readout   = GlobalReduce(reduce_op="sum")
head      = Linear(hidden_channels, num_classes)

# Forward pass: MP -> Pool -> MP -> Readout -> Head.
x   = F.relu(conv_pre(batch.x, batch.edge_index, batch.edge_weight))
out = pooler(x=x, adj=batch.edge_index, batch=batch.batch)
if use_dense_branch:
    x_next    = F.relu(conv_post(out.x, out.edge_index, mask=out.mask))
    graph_emb = readout(x_next, mask=out.mask)
else:
    x_next    = F.relu(conv_post(out.x, out.edge_index, out.edge_weight))
    graph_emb = readout(x_next, batch=out.batch)
logits = head(graph_emb)
loss   = task_loss(logits, batch.y)
if out.loss is not None:
    loss = loss + sum(out.get_loss_value())
\end{lstlisting}

%% file: listings/node-level.tex
\begin{lstlisting}[language=Python, caption={Node-classification pipeline with pooling, bottleneck message passing, and lifting.}, label={lst:node-class-pool-lift}, style=tgpcode]
self.conv_enc = GINConv(in_channels, hidden_channels)
self.pooler = get_pooler(
    "mincut",
    in_channels=hidden_channels,
    k=num_nodes // 20,
    batched=True,
    sparse_output=False,
    cache_preprocessing=True,
)

self.conv_pool = DenseGINConv(hidden_channels, hidden_channels // 2)
self.conv_dec = DenseGINConv(hidden_channels // 2, num_classes)

# Forward pass
x = F.relu(self.conv_enc(x, edge_index, edge_weight))
out = self.pooler(x=x, adj=edge_index, edge_weight=edge_weight, batch=batch)
x_pool = F.relu(self.conv_pool(out.x, out.edge_index, mask=out.mask))

# Important: use lifting=True to switch to the lifting pass.
x_lift = self.pooler(x=x_pool, so=out.so, lifting=True,
                     batch=batch, batch_pooled=out.batch)
adj_dense = self.pooler.preprocessing_cache
logits = self.conv_dec(x_lift, adj_dense)

loss = F.nll_loss(F.log_softmax(logits, dim=-1)[train_mask], y[train_mask])
if out.loss is not None:
    loss = loss + sum(out.get_loss_value())
\end{lstlisting}

%% file: listings/clustering.tex
\begin{lstlisting}[language=Python, caption={Unsupervised vertex clustering: ARMA encoder, dense pooler, and optimization on auxiliary losses only.}, label={lst:clustering-unsupervised}, style=tgpcode]
from torch.nn import ELU, ModuleList
from torch_geometric.nn import ARMAConv
from tgp.poolers import get_pooler

embed_dim = 32  # set to benchmark width (see clustering results)

self.mp = ModuleList([
    ARMAConv(in_channels, embed_dim, num_layers=2), ELU(inplace=True),
    ARMAConv(embed_dim, embed_dim, num_layers=2), ELU(inplace=True),
])
self.pooler = get_pooler(
    "mincut",
    in_channels=embed_dim,
    k=num_clusters,
    cache_preprocessing=True,
)

def forward(self, x, edge_index, edge_weight):
    for i in range(0, len(self.mp), 2):
        x = self.mp[i](x, edge_index, edge_weight)
        x = self.mp[i + 1](x)
    out = self.pooler(x=x, adj=edge_index, edge_weight=edge_weight)
    aux_loss = sum(out.get_loss_value())
    return out.so.s, aux_loss
\end{lstlisting}

%% file: listings/aggr-reduce.tex
\begin{lstlisting}[language=Python, caption={Sparse \gls{topk} graph classification with \texttt{AggrReduce} and \texttt{GlobalReduce}.}, label={lst:aggr-reduce}, style=tgpcode]
from tgp.poolers import get_pooler
from tgp.reduce import AggrReduce, GlobalReduce, get_aggr

pooler = get_pooler("topk", in_channels=hidden_channels, ratio=0.25)

# Swap pooler's reduce stage with a PyG-backed aggregator.
aggr_name = "set2set"
aggr_kwargs = {"in_channels": hidden_channels, "processing_steps": 3}
pooler.reducer = AggrReduce(get_aggr(aggr_name, **aggr_kwargs))

# Use the same aggregation family at graph readout.
readout = GlobalReduce(reduce_op=aggr_name, **aggr_kwargs)

out = pooler(x=batch.x, adj=batch.edge_index,
             edge_weight=getattr(batch, "edge_weight", None),
             batch=batch.batch)

x_next = sparse_conv(out.x, out.edge_index, out.edge_weight)
graph_emb = readout(x_next, batch=out.batch)

logits = head(graph_emb)
loss = task_loss(logits, batch.y)
if out.loss is not None:
    loss = loss + sum(out.get_loss_value())
\end{lstlisting}

%% file: tables/flags.tex
\begin{table}[htbp]
       \centering
       \small
       \setlength{\tabcolsep}{5pt}
       \renewcommand{\arraystretch}{1.08}
       \begin{tabularx}{\linewidth}{>{\raggedright\arraybackslash}p{0.20\linewidth}>{\raggedright\arraybackslash}p{0.30\linewidth}X}
              \toprule
              \textbf{Flag}               & \textbf{Meaning}                      & \textbf{Typical usage in model code}                                                                                             \\
              \midrule
              \texttt{is\_dense}          & Pooler uses dense assignment          & Select dense execution settings (for example \texttt{batched} and \texttt{sparse\_output}) and dense-compatible message passing. \\
              \texttt{has\_loss}          & Pooler returns auxiliary losses       & Read losses from \texttt{PoolingOutput} and add them to task loss in the training loop.                                          \\
              \texttt{is\_precoarsenable} & Coarsening can be precomputed offline & Route to \texttt{PreCoarsening} pipelines and skip repeated online coarsening.                                                   \\
              \bottomrule
       \end{tabularx}
       \caption{Compatibility flags in \texttt{tgp}; they let model code choose the correct execution path without branching on specific pooler names.}
       \label{tab:flags-workflow}
\end{table}

%% file: listings/topk.tex
\begin{lstlisting}[language=Python, caption={Sketched implementation of \texttt{TopkPooling}, showcasing the modular design. The \texttt{\_\_init\_\_} defines the \gls{srcl} components of the pooler, which are then called in the \texttt{forward} pass.}, label={lst:topk_implementation}, style=tgpcode]
class TopkPooling(SRCPooling):
  def __init__(self, in_channels, ratio, **kwargs):
    super().__init__(
        selector=TopkSelect(in_channels=in_channels, ratio=ratio, ...), 
        reducer=BaseReduce(),
        connector=SparseConnect(), 
        lifter=BaseLift())

  def forward(self, x, edge_index, batch=None):
    # 1. Select nodes via the TopkSelect component
    select_out = self.select(x=x, batch=batch)
    
    # 2. Reduce features via the BaseReduce component
    x_pooled, batch_pooled = self.reduce(x=x, so=select_out, batch=batch)
    
    # 3. Connect graph via the SparseConnect component
    edge_index_pooled, _ = self.connect(so=select_out, edge_index=edge_index)
    
    # 4. Return standardized output
    return PoolingOutput(x=x_pooled, edge_index=edge_index_pooled,
        batch=batch_pooled, so=select_out)
\end{lstlisting}

%% file: listings/swap.tex
\begin{lstlisting}[language=Python, caption={Instantiating a custom pooling layer by overriding the connector component.}, label={lst:custom_layer}, style=tgpcode]
from tgp.poolers import get_pooler
from tgp.connect import KronConnect

# 1. Instantiate a standard TopK pooler
pooler = get_pooler("topk", in_channels=64, ratio=0.5)

# 2. Hot-swap the connector component
# Replace the standard connect operation with the one based on Kron connect
pooler.connector = KronConnect()
\end{lstlisting}

%% file: tables/poolers.tex
\begin{table*}[htbp]
\caption{User-facing characteristics of the pooling operators implemented in \gls{tgp}. The ``Dense execution'' column refers to input handling for dense poolers: ``both'' means that the operator supports the padded batched path and the unbatched path; ``unbatched'' means that only the unbatched dense path is available.}
\label{tab:pooler_features}
\centering
\scriptsize
\resizebox{\textwidth}{!}{
\begin{tabular}{lccccc}
\cmidrule[1.5pt]{1-6}
\textbf{Operator} & \textbf{Assignment} & \textbf{Trainable} & \textbf{Aux. loss} & \textbf{Dense execution} & \textbf{Pre-coarsening} \\
\midrule
\gls{asap} & sparse & \checkmark & & n/a & \\
\gls{acc} & dense & \checkmark & \checkmark & both & \\
\gls{bnpool} & dense & \checkmark & \checkmark & both & \\
\gls{dmon} & dense & \checkmark & \checkmark & both & \\
\gls{diff} & dense & \checkmark & \checkmark & both & \\
\gls{ecpool} & sparse & \checkmark & & n/a & \\
\gls{eigen} & dense & & & unbatched & \checkmark \\
\gls{graclus} & sparse & & & n/a & \checkmark \\
\gls{hosc} & dense & \checkmark & \checkmark & both & \\
\gls{jbgnn} & dense & \checkmark & \checkmark & both & \\
\gls{kmis} & sparse & & & n/a & \checkmark \\
\gls{lapool} & dense & \checkmark & & both & \\
\gls{maxcut} & sparse & \checkmark & \checkmark & n/a & \\
\gls{mincut} & dense & \checkmark & \checkmark & both & \\
\gls{ndp} & sparse & & & n/a & \checkmark \\
\gls{nmf} & dense & & & unbatched & \checkmark \\
\gls{pan} & sparse & \checkmark & & n/a & \\
\gls{sag} & sparse & \checkmark & & n/a & \\
\gls{sep} & sparse & & & n/a & \checkmark \\
\gls{topk} & sparse & \checkmark & & n/a & \\
\cmidrule[1.5pt]{1-6}
\end{tabular}
}
\end{table*}

%% file: tables/datasets/clustering.tex
\begin{table}[htbp]
\caption{Details of the vertex-clustering datasets.}
\label{tab:clustering_datasets}
\centering
\begin{tabular}{lcccc}
\cmidrule[1.5pt]{1-5}
\textbf{Dataset} & \textbf{\#Vertices} & \textbf{\#Edges} & \textbf{\#Vertex attr.} & \textbf{\#Classes} \\
\midrule
Community & 400 & 5,904 & 2 & 5 \\
Cora & 2,708 & 10,556 & 1,433 & 7 \\
Citeseer & 3,327 & 9,104 & 3,703 & 6 \\
Pubmed & 19,717 & 88,648 & 500 & 3 \\
DBLP & 17,716 & 105,734 & 1,639 & 4 \\
\cmidrule[1.5pt]{1-5}
\end{tabular}
\end{table}

%% file: tables/datasets/node-level.tex
\begin{table}[htbp]
\caption{Statistics of the heterophilic node-classification datasets.}
\label{tab:node_class_datasets_app}
\centering
\begin{tabular}{lcccc}
\cmidrule[1.5pt]{1-5}
\textbf{Dataset} & \textbf{\# Nodes} & \textbf{\# Edges} & \textbf{\# Classes} & $\boldsymbol{h(\mathcal{G})}$ \\
\midrule
Roman-Empire & 22,662 & 32,927 & 18 & 0.021 \\
Amazon-Ratings & 24,492 & 93,050 & 5 & 0.127 \\
Minesweeper & 10,000 & 39,402 & 2 & 0.009 \\
Tolokers & 11,758 & 519,000 & 2 & 0.180 \\
Questions & 48,921 & 153,540 & 2 & 0.079 \\
\cmidrule[1.5pt]{1-5}
\end{tabular}
\end{table}

%% file: tables/datasets/graph-level.tex
\begin{table}[htbp]
\caption{Details of graph-level datasets.}
\label{tab:graph_class_datasets_app}
\centering
\begin{tabular}{l c c c c}
\cmidrule[1.5pt]{1-5}
\textbf{Dataset} & \textbf{\# Graphs} & \textbf{\# Classes} & \textbf{Avg. \# Nodes} & \textbf{Avg. \# Edges} \\
\midrule
NCI1 & 4,110 & 2 & 29.87 & 64.60 \\
Reddit-binary & 2,000 & 2 & 429.63 & 497.75 \\
EXPWL1 & 3,000 & 2 & 76.96 & 186.46 \\
Multipartite & 5,000 & 10 & 99.79 & 4,477.43 \\
GCB-H & 1,800 & 3 & 148.32 & 572.32 \\
CSBM & 1,000 & 2 & 100.00 & 984.87 \\
\cmidrule[1.5pt]{1-5}
\end{tabular}
\end{table}

%% file: tables/results/clustering.tex
\bgroup
\setlength\tabcolsep{1.0em} 
\begin{table*}[htbp]
\caption{NMI by different poolers on node clustering.}
\label{tab:clustering_results}
\centering
{\small
\begin{tabular}{l|ccccc|c}
\cmidrule[1.5pt]{1-7}
\textbf{Pooler} & \textbf{CiteSeer} & \textbf{Community} & \textbf{Cora} & \textbf{DBLP} & \textbf{PubMed} & \textbf{Score} \\
\cmidrule[1pt]{1-7}
\gls{acc} & \textbf{27 {\tiny $\pm$ 3}} & \underline{95 {\tiny $\pm$ 4}} & \textbf{42 {\tiny $\pm$ 3}} & 28 {\tiny $\pm$ 3} & 19 {\tiny $\pm$ 4} & \underline{5} \\
\gls{diff} & 15 {\tiny $\pm$ 3} & 61 {\tiny $\pm$ 2} & 29 {\tiny $\pm$ 2} & 16 {\tiny $\pm$ 3} & 10 {\tiny $\pm$ 2} & 0 \\
\gls{dmon} & \underline{23 {\tiny $\pm$ 1}} & \textbf{99 {\tiny $\pm$ 1}} & \underline{37 {\tiny $\pm$ 2}} & 27 {\tiny $\pm$ 4} & \textbf{23 {\tiny $\pm$ 6}} & \textbf{6} \\
\gls{hosc} & 18 {\tiny $\pm$ 4} & 91 {\tiny $\pm$ 6} & 30 {\tiny $\pm$ 2} & \textbf{34 {\tiny $\pm$ 1}} & 19 {\tiny $\pm$ 3} & 2 \\
\gls{jbgnn} & 18 {\tiny $\pm$ 5} & 94 {\tiny $\pm$ 8} & 23 {\tiny $\pm$ 4} & 18 {\tiny $\pm$ 9} & 18 {\tiny $\pm$ 3} & 0 \\
\gls{mincut} & 22 {\tiny $\pm$ 4} & 92 {\tiny $\pm$ 1} & 37 {\tiny $\pm$ 5} & \underline{33 {\tiny $\pm$ 4}} & \underline{20 {\tiny $\pm$ 1}} & 2 \\
\cmidrule[1.5pt]{1-7}
\end{tabular}
}
\end{table*}
\egroup

%% file: tables/results/node-level.tex
\bgroup
\setlength\tabcolsep{1.0em} 
\begin{table*}[htbp]
\caption{Test accuracy (Roman-Empire, Amazon-Ratings) and AUROC (Minesweeper, Tolokers, Questions) by different poolers on node classification.}
\label{tab:node_classification_results}
\centering
{\small
\resizebox{0.98\textwidth}{!}{%
\begin{tabular}{l|ccccc|c}
\cmidrule[1.5pt]{1-7}
\textbf{Pooler} & \textbf{\shortstack{Amazon-\\Ratings}} & \textbf{Minesweeper} & \textbf{Questions} & \textbf{\shortstack{Roman-\\Empire}} & \textbf{Tolokers} & \textbf{Score} \\
\cmidrule[1pt]{1-7}
\gls{asap} & 40 {\tiny $\pm$ 2} & \textbf{78 {\tiny $\pm$ 5}} & 62 {\tiny $\pm$ 6} & 27 {\tiny $\pm$ 1} & \underline{79 {\tiny $\pm$ 1}} & 3 \\
\gls{kmis} & \textbf{44 {\tiny $\pm$ 1}} & \underline{71 {\tiny $\pm$ 1}} & \textbf{66 {\tiny $\pm$ 1}} & \underline{32 {\tiny $\pm$ 1}} & 77 {\tiny $\pm$ 5} & \textbf{6} \\
\gls{maxcut} & 42 {\tiny $\pm$ 1} & 63 {\tiny $\pm$ 4} & \underline{65 {\tiny $\pm$ 2}} & \textbf{36 {\tiny $\pm$ 2}} & \textbf{80 {\tiny $\pm$ 1}} & \underline{5} \\
\gls{ndp} & \underline{43 {\tiny $\pm$ 1}} & 62 {\tiny $\pm$ 1} & 59 {\tiny $\pm$ 6} & 23 {\tiny $\pm$ 1} & 71 {\tiny $\pm$ 6} & 1 \\
\gls{topk} & 38 {\tiny $\pm$ 1} & 66 {\tiny $\pm$ 2} & 62 {\tiny $\pm$ 3} & 21 {\tiny $\pm$ 2} & 70 {\tiny $\pm$ 3} & 0 \\
\cmidrule[1.5pt]{1-7}
\end{tabular}
}
}
\end{table*}
\egroup

%% file: tables/results/graph-level.tex
\bgroup
\setlength\tabcolsep{1.0em} 
\begin{table*}[htbp]
\caption{Test accuracy by different poolers on graph-level classification.}
\label{tab:graph_classification_results}
\centering
\resizebox{0.98\textwidth}{!}{%
\begin{tabular}{l|cccccc|c}
\cmidrule[1.5pt]{1-8}
\textbf{Pooler} & \textbf{NCI1} & \textbf{GCB-H} & \textbf{CSBM} & \textbf{EXPWL1} & \textbf{Multipartite} & \textbf{\shortstack{Reddit-\\binary}} & \textbf{Score} \\
\cmidrule[1pt]{1-8}
\gls{acc} & 77 {\tiny $\pm$ 2} & 71 {\tiny $\pm$ 1} & 67 {\tiny $\pm$ 4} & 92 {\tiny $\pm$ 1} & 61 {\tiny $\pm$ 3} & 90 {\tiny $\pm$ 3} & 0 \\
\gls{asap} & 76 {\tiny $\pm$ 2} & 71 {\tiny $\pm$ 4} & 79 {\tiny $\pm$ 5} & \textit{99 {\tiny $\pm$ 0}} & 12 {\tiny $\pm$ 3} & 88 {\tiny $\pm$ 3} & 1 \\
\gls{bnpool} & \textbf{78 {\tiny $\pm$ 2}} & 71 {\tiny $\pm$ 2} & 69 {\tiny $\pm$ 4} & 93 {\tiny $\pm$ 1} & \underline{65 {\tiny $\pm$ 2}} & \textit{91 {\tiny $\pm$ 2}} & \underline{6} \\
\gls{diff} & \textit{77 {\tiny $\pm$ 3}} & 67 {\tiny $\pm$ 2} & 55 {\tiny $\pm$ 7} & 96 {\tiny $\pm$ 1} & 59 {\tiny $\pm$ 2} & 90 {\tiny $\pm$ 2} & 1 \\
\gls{dmon} & 77 {\tiny $\pm$ 2} & 71 {\tiny $\pm$ 1} & 67 {\tiny $\pm$ 3} & 93 {\tiny $\pm$ 1} & 61 {\tiny $\pm$ 2} & 91 {\tiny $\pm$ 1} & 0 \\
\gls{ecpool} & \textbf{78 {\tiny $\pm$ 2}} & \underline{73 {\tiny $\pm$ 2}} & \textbf{99 {\tiny $\pm$ 1}} & 99 {\tiny $\pm$ 1} & 10 {\tiny $\pm$ 1} & 90 {\tiny $\pm$ 2} & \textbf{8} \\
\gls{eigen} & 69 {\tiny $\pm$ 2} & 66 {\tiny $\pm$ 2} & 68 {\tiny $\pm$ 5} & 98 {\tiny $\pm$ 1} & \textbf{66 {\tiny $\pm$ 1}} & 90 {\tiny $\pm$ 3} & 3 \\
\gls{graclus} & 76 {\tiny $\pm$ 3} & 72 {\tiny $\pm$ 1} & 96 {\tiny $\pm$ 2} & 99 {\tiny $\pm$ 1} & 10 {\tiny $\pm$ 0} & 90 {\tiny $\pm$ 2} & 0 \\
\gls{hosc} & \textbf{78 {\tiny $\pm$ 2}} & 71 {\tiny $\pm$ 1} & 67 {\tiny $\pm$ 3} & 95 {\tiny $\pm$ 3} & 62 {\tiny $\pm$ 2} & \underline{91 {\tiny $\pm$ 1}} & \textit{5} \\
\gls{jbgnn} & \textbf{78 {\tiny $\pm$ 2}} & 72 {\tiny $\pm$ 2} & 63 {\tiny $\pm$ 3} & 99 {\tiny $\pm$ 1} & 58 {\tiny $\pm$ 3} & \textbf{91 {\tiny $\pm$ 2}} & \underline{6} \\
\gls{kmis} & 77 {\tiny $\pm$ 2} & \textit{72 {\tiny $\pm$ 1}} & \textbf{99 {\tiny $\pm$ 1}} & \textbf{100 {\tiny $\pm$ 0}} & \textit{63 {\tiny $\pm$ 2}} & 91 {\tiny $\pm$ 3} & \textbf{8} \\
\gls{lapool} & \underline{77 {\tiny $\pm$ 2}} & \textit{72 {\tiny $\pm$ 1}} & 69 {\tiny $\pm$ 5} & 96 {\tiny $\pm$ 2} & 61 {\tiny $\pm$ 2} & 90 {\tiny $\pm$ 1} & 3 \\
\gls{maxcut} & 77 {\tiny $\pm$ 2} & 70 {\tiny $\pm$ 3} & \underline{98 {\tiny $\pm$ 1}} & \textbf{100 {\tiny $\pm$ 0}} & 60 {\tiny $\pm$ 2} & 87 {\tiny $\pm$ 3} & \textit{5} \\
\gls{mincut} & \textbf{78 {\tiny $\pm$ 2}} & 71 {\tiny $\pm$ 1} & 67 {\tiny $\pm$ 5} & 92 {\tiny $\pm$ 1} & 61 {\tiny $\pm$ 2} & 90 {\tiny $\pm$ 2} & 3 \\
\gls{ndp} & 76 {\tiny $\pm$ 2} & 72 {\tiny $\pm$ 2} & \textbf{99 {\tiny $\pm$ 1}} & 98 {\tiny $\pm$ 1} & 13 {\tiny $\pm$ 9} & 87 {\tiny $\pm$ 2} & 3 \\
\gls{pan} & 71 {\tiny $\pm$ 3} & 48 {\tiny $\pm$ 12} & \textit{97 {\tiny $\pm$ 2}} & 74 {\tiny $\pm$ 6} & 10 {\tiny $\pm$ 0} & 87 {\tiny $\pm$ 5} & 1 \\
\gls{sag} & 77 {\tiny $\pm$ 3} & 65 {\tiny $\pm$ 5} & 75 {\tiny $\pm$ 8} & 93 {\tiny $\pm$ 5} & 51 {\tiny $\pm$ 15} & 87 {\tiny $\pm$ 3} & 0 \\
\gls{sep} & 77 {\tiny $\pm$ 2} & \textbf{73 {\tiny $\pm$ 1}} & 69 {\tiny $\pm$ 4} & \underline{100 {\tiny $\pm$ 1}} & 62 {\tiny $\pm$ 2} & 89 {\tiny $\pm$ 3} & \textit{5} \\
\gls{topk} & 75 {\tiny $\pm$ 2} & 68 {\tiny $\pm$ 4} & 60 {\tiny $\pm$ 9} & 91 {\tiny $\pm$ 3} & 55 {\tiny $\pm$ 3} & 86 {\tiny $\pm$ 3} & 0 \\
\cmidrule[1.5pt]{1-8}
\end{tabular}
}
\end{table*}
\egroup